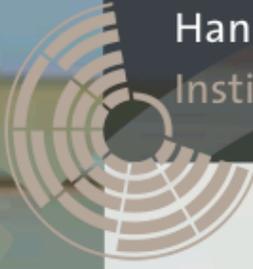

# Hanse-Wissenschaftskolleg
## Institute for Advanced Study

# 2nd Symposium on Problem-Solving, Creativity and Spatial Reasoning in Cognitive Systems

Delmenhorst, July 20 - 21, 2017

**Organizers:**
Dr.-Ing. Zoe Falomir
Dr. Dr. Ana-Maria Olteţeanu
Universität Bremen

Program

**Venue:**

Hanse-Wissenschaftskolleg
Institute for Advanced Study
Lehmkuhlenbusch 4
27753 Delmenhorst
Germany
www.h-w-k.de

Proceedings of the

**ProSocrates'17 Symposium:**
**Problem-solving, Creativity and Spatial Reasoning in Cognitive Systems**

Delmenhorst, Germany, 20-21 July 2017

Edited by

Ana-Maria Olteteanu
Bremen Spatial Cognition Centre (BSCC)
University of Bremen, Germany

and

Zoe Falomir Llansola
Bremen Spatial Cognition Centre (BSCC)
University of Bremen, Germany

This book contains the accepted papers at *ProSocrates'17 Symposium: Problem-solving, Creativity and Spatial Reasoning in Cognitive Systems*[1]. ProSocrates'17 symposium was held at the Hansewissenschaftkolleg (HWK) of Advanced Studies in Delmenhorst, 20-21 July 2017. This was the second edition of this symposium which tries to bring together researchers interested in spatial reasoning, problem solving and creativity.

These ProSocrates'17 proceedings contains 6 accepted papers that were presented at the symposium. Each submitted paper was reviewed by two/three program committee members. Moreover, there were 6 invited talks whose abstracts are also included in these proceedings.

**Acknowledgements**

The funding from the Hansewissenschaftkolleg (HWK) of Advanced Studies[2] is gratefully acknowledged.

The support from the projects *Cognitive Qualitative Descriptions and Applications*[3] (CogQDA) funded by the University of Bremen and *Creative Problem Solving in Cognitive Systems*[4] (CreaCogs) funded by DFG are acknowledged. The collaboration from our colleagues at Bremen Spatial Cognition Centre (BSCC) is also acknowledged.

We would like also to thank the members of the Scientific Committee for their valuable work during the reviewing process and the additional reviews.

We also thank Easychair, which was used to manage paper submissions and reviewing the proceedings, and CEUR Workshop Proceedings (CEUR-WS.org) for indexing these proceedings.

Zoe Falomir Llansola
Ana-Maria Olteteanu
ProSocrates'17 Chairs
July 2017

---

[1] ProSocrates'17: https://prosocrates.wordpress.com/
[2] HWK: http://www.h-w-k.de/
[3] CogQDA: https://sites.google.com/site/cogqda/
[4] CreaCogs: http://creacogcomp.com/

# Table of Contents: ProSocrates'17

# Reasoning at a Distance by Way of Conceptual Metaphors and Blends

Marco Schorlemmer
Artificial Intelligence Research Institute (IIIA-CSIC)
Barcelona, Spain
marco@iiia.csic.es

## Abstract

Cognitive scientists of the embodied cognition tradition have been providing evidence that a large part of our creative reasoning and problem-solving processes are carried out by means of conceptual metaphor and blending, grounded on our bodily experience with the world. In this talk I shall aim at fleshing out a mathematical model that has been proposed in the last decades for expressing and exploring conceptual metaphor and blending with greater precision than has previously been done. In particular, I shall focus on the notion of aptness of a metaphor or blend and on the validity of metaphorical entailment. Towards this end, I shall use a generalisation of the category-theoretic notion of colimit for modelling conceptual metaphor and blending in combination with the idea of reasoning at a distance as modelled in the Barwise-Seligman theory of information flow. I shall illustrate the adequacy of the proposed model with an example of creative reasoning about space and time for solving a classical brain-teaser. Furthermore, I shall argue for the potential applicability of such mathematical model for ontology engineering, computational creativity, and problem-solving in general.

# Symbolic models and computational properties of constructive reasoning in cognition and creativity

Tarek R. Besold
Digital Media Lab
Center for Computing and Communication Technologies (TZI)
University of Bremen, Germany
`tarek.besold@uni-bremen.de`

## Abstract

Analogy is one of the most studied forms of non-classical reasoning working across different domains, usually taken to play a crucial role in creative thought and problem-solving. In the first part of the talk, I will introduce general principles of computational analogy models (relying on a generalisation-based approach to analogy-making). We will then have a closer look at Heuristic-Driven Theory Projection (HDTP) as an example for a theoretical framework and implemented system: HDTP computes analogical relations and inferences for domains which are represented using many-sorted first-order logic languages, applying a restricted form of higher-order anti-unification for finding shared structural elements common to both domains. The presentation of the framework will be followed by a few reflections on the cognitive plausibility of the approach motivated by theoretical complexity and tractability considerations. In the second part I will touch upon several applications of HDTP to modeling important cognitive capacities, including concept blending processes as current hot topic in Cognitive Science.

# Creating and rating harmonic colour palettes for a given style

Lledó Museros
Universitat Jaume I
Castellón, Spain
`museros@uji.es`

## Abstract

Colour, and more specifically, colour harmony has an important role in creativity and design. During the talk a qualitative colour theory, and the operations to create harmonic colour palettes, will be presented. Then the process to classify these palettes as a life-style, as for instance casual, romantic, elegant, and so on will be introduced. Moreover, the palettes generated can be rated in function of the taste of the people by using social data. On the other hand, how these information can be used to characterise images, and also to give emotional descriptors to the images will be presented.

# Creative Support Companions: Some Ideas

Ken Forbus
Walter P. Murphy Professor of Computer Science and
Professor of Education at
Northwestern University
`forbus@northwestern.edu`

## Abstract

An exciting opportunity for AI is the development of intelligent assistants that, working with people, enable them to do far more than they can alone. What would that mean for creative activities? This talk explores some ideas for using the Companion cognitive architecture to create software collaborators that support creative work. Companions include human-like analogical processing, facilities for natural language and sketch understanding, and rich relational representations that capture aspects of human visual, spatial, and conceptual knowledge. For supporting creative activities, this should enable them to (1) help suggest and explore cross-domain analogies, (2) interact via natural modalities, providing higher communication bandwidth and reducing friction compared to software tools, and (3) adapt to their human partners over time, building up a portfolio of joint work that can be drawn upon in future efforts.

# Modeling visual problem-solving as analogical reasoning

Andrew Lovett
Northwestern University Chicago, US
`andrew@cs.northwestern.edu`

## Abstract

Visual problem-solving tasks are powerful tools for evaluating intelligence and creative thinking in humans. For example, the Ravens Progressive Matrices is one of the best single-test predictors of a persons spatial, verbal, and mathematical ability. To better understand the skills that allow people to succeed at problem-solving, I have developed a computational model. The model builds on the claim that analogical reasoning lies at the heart of visual problem-solving. Images are compared via structure-mapping, aligning the common relational structure in two images to identify commonalities and differences. These commonalities or differences can themselves be reified and used as the input for future comparisons. When images fail to align, the model re-represents them to facilitate the comparison. In this talk, I describe what the model has taught me, in terms of the challenges faced during problem-solving and the skills that can be used to overcome those challenges.

# Thinking Like A Child: The Role of Surface Similarities in Stimulating Creativity

Bipin Indurkhya
Computer Science and Cognitive Science Dep.
Jagiellonian University
Kraków, Poland
`bipin.indurkhya@uj.edu.pl`

## Abstract

An oft-touted mantra for creativity is: think like a child. We focus on one particular aspect of child-like thinking here, namely surface similarities. Developmental psychology has convincingly demonstrated, time and again, that younger children use surface similarities for categorization and related tasks; only as they grow older they start to consider functional and structural similarities. We consider examples of puzzles, research on creative problem solving, and two of our recent empirical studies to demonstrate how surface similarities can stimulate creative thinking. We examine the implications of this approach for designing creativity-support systems.

# Towards finer-grained interaction with a Poetry Generator

Hugo Gonçalo Oliveira
`hroliv@dei.uc.pt`

Tiago Mendes
`tjmendes@student.dei.uc.pt`

Ana Boavida
`aboavida@dei.uc.pt`

CISUC, Department of Informatics Engineering
University of Coimbra, Portugal

## Abstract

PoeTryMe is a poetry generation system that produces poems autonomously, from a set of initial parameters. After using a simplified version of this system, creative writers and other interested people identified some issues and expressed their wish to make changes in the resulting poems or to interact with the system and take part in the creative process. This paper illustrates some of the issues on generated poems and reports on a recent effort towards providing alternative ways of using PoeTryMe and to meet the previous suggestions. Some functionalities were made available via a web API, which can now be exploited by other systems. Those include the generation of single lines, or the retrieval of related words, where additional constrains can be made on the number of syllables, sentiment or rhyme. On the top of this API, a co-creative interface has been developed. It enables users to start from scratch, from an existing poem, or to generate a new draft.

## 1 Introduction

PoeTryMe [6] is a platform for automated poetry generation. So far, it has been adapted to produce poetry in different forms, languages and from different stimuli. PoeTryMe generates a poem from a set of initial parameters, such as the poetry form, the language, a set of seeds or a surprise factor. An implemented generation strategy is then followed to select the most suitable lines to fill the form, generated with the help of a semantic network and a generation grammar. The result is a sequence of semantically-coherent lines, using the seeds or words related to them, grouped according to the given poetry form, with the corresponding number of syllables, and often with rhymes. This is done in an autonomous fashion, with no additional user interaction nor input besides the initial parameters.

Yet, after using a simplified version of PoeTryMe, several users, including musicians and poets, expressed their interest to be more involved in PoeTryMe's creation process. Some even confessed to having generated several poems, keeping only some of the best lines, and created a new improved poem from this manual procedure. Although the obtained results are generally ok, especially if generation goes through more iterations in the search for the most suitable lines, there are always aspects that may be improved, not to mention that assessing the quality of poetry is a subjective task and may diverge from user to user.

In order to meet the users feedback, we decided to provide direct access to some of the main functionalities required for poetry generation, through a web API. This was possible due to the modular architecture of PoeTryMe, which already decouples the process of poetry generation in several tasks, performed independently.

Functionalities now available through the API include the generation of single lines from a seed, with a target number of syllables, or the retrieval of words related to another, where additional constraints can be made on the number of syllables, sentiment or rhyme. To some extent, this is in line with the vision of a Creative Web, where creative tools are deployed as web services [20], and enables third-party applications to interact with PoeTryMe, not only for the generation of a single poem, but also to customise when each of the available functionalities is performed, possibly combining them in alternative ways.

One of such applications is Co-PoeTryMe, a web-based tool that enables to use PoeTryMe co-creatively. Users may start from scratch, with their own words and lines, they can import an existing poem, or they can generate a newl draft poem, which should help them to overcome the 'blank page syndrome'[1] [14]. In any case, lines and words may then be switched or replaced with new ones, following certain user-defined constraints, hopefully towards a better poem, more aligned with their intentions.

The remaining of this paper starts with a brief introduction to related topics, including poetry generation, co-creativity, poetry and co-creativity, and finally creativity-support interfaces. After this, PoeTryMe is overviewed, followed by an enumeration of some identified limitations, with illustrative examples. Together with the user feedback, the previous limitations motivate most functionalities that were made available, described next. Before concluding, Co-PoeTryMe is revealed and its main features are briefly described.

## 2 Related Work

Poetry generation is a popular task in the scope of Computational Creativity [3]. During the last 15 years, several systems for this purpose were developed, using a wide range of artificial intelligence techniques (e.g. [4, 17, 2, 19, 6, 18, 16], see [10] for a short survey). The majority of those systems generates poetry autonomously, from initial stimuli, either provided by the user or triggered by some event, and outputs the results without any human intervention during the generative process.

A different kind of systems enable the collaboration of humans and computational agents in the production of creative artefacts. This can be done through alternating co-creativity, when the computer performs exactly the same tasks as the human, though in different turns; or through task-divided co-creativity, when the tasks performed by the computer are different than those by the human [15]. For instance, in the poetry domain, some tasks should be easier for a computer program – e.g., finding words that rhyme or with a certain number of syllables, which, to some extent, can be made with the help of a set of rules and / or a pronunciation lexicon – , while other tasks would suit the human better – e.g., evaluating the aesthetics, which can be very subjective; or conveying a non-trivial meaning, which might require access to much knowledge organised for this purpose.

We identified existing computational systems that enable the collaboration between man and machine in the production of poems [14, 16, 21], though with different interfaces and available functionalities. The Poetry Machine [14] is presented as an interactive tool developed on the top of a poetry generation system [19]. From a set of user-given keywords, the system generates initial draft lines, which may be further changed by the user or by the system, in equal turns. During the creation process, the user may ask for additional lines or words, while the new fragments produced by the system will automatically adapt to the user's modifications. jGnoetry is a web application[2] that enables the generation of poems from a collection of texts provided by the user, who may also select the poetry form from a list of predefined forms or set its own through sequences of syllables organised in lines. The resulting poem is presented in such a way that the user may select words to keep in further iterations and generate new text for the unselected words. Deep Beat [16] is a rap lyrics generation system with a web interface[3] where the users can write some lines, provide a set of words to appear, and ask the system for the suggestion of new lines, possibly rhyming. Each suggested line appears after a picture of its original author, because they were collected from human-created rap lyrics. Another lyrics-writing supporting system is LyriSys [21], where the user sets a musical structure, chooses a topic for each block or writes part of the lyrics, and the system generates lyrics for the remaining blocks, following the given structure and within the selected topic. After this, users can still revise the lines they are not happy with, possibly replacing them with alternative candidates.

From a similar perspective, Misztal and Indurkhya's [18] poetry generation system could also be seen as co-creative, but the collaboration does not involve humans, only computational agents with different expertises.

---

[1] The 'blank page syndrome' refers to when a writer opens a blank page and cannot or takes too long to start writing because there are no words on the page
[2] http://www.eddeaddad.net/jgnoetry/
[3] http://deepbeat.org/

More precisely, agents with different responsibilities, such as dealing with emotion, word generation, poetic aspects, or selecting the best solutions, interact in a common workspace (blackboard), towards the production of a poem.

A different kind of creativity platform is not co-creative nor creative on its own, but allows users to combine different more or less creative services through a web interface, in the development of novel creative workflows, which can be tested right away. Such platforms include ConCreTeFlows [22] and FloWr [1] and have been used for poetry generation, among other tasks.

In order to be integrated in such platforms, creative systems must somehow be in line with the vision of a Creative Web, where creative tools are deployed as web services [20] that may be used by third-party applications. Concerning poetry generation, the previous vision is further discussed by Gervás [5], who argues that abstractions of the various functionalities involved in a poetry generation system should be available as services that may be later invoked by other systems. This enables the development of different systems that would, nevertheless, share some modules, thus requiring less effort to develop. As for the co-creative systems, they must communicate with their user interfaces and allow to run different steps involved in poetry generation at will, instead of always going through the full process. So, even if these steps are not always available as services, modularity should be present and, in most cases, each module could potentially become available as an independent service.

## 3 PoeTryMe and its Modular Architecture

PoeTryMe [6] is a computational platform, originally designed to test different settings in the process of poetry generation, with a focus on the exploited knowledge resources, which could be indirectly assessed this way. For this reason, PoeTryMe has a modular architecture that enables not only the independent development of each module, but also to test different settings of input parameters and to study their impact in the resulting poems, with reduced effort. PoeTryMe's architecture has two core modules: the Line Generator produces semantically-coherent lines with the help of a semantic network and a generation grammar, with textual patterns for rendering semantic relations, given their type; and the Generation Strategy retrieves lines on a semantic domain, from the Line Generator, and selects which will be used in the poem. Additional modules can be seen as complementary. A more extensive description of this architecture is found elsewhere [11].

Among other parameters, users may define the rules of a generation grammar, the underlying semantic network, the poetry form, the set of seed words, the polarity lexicon and the transmitted sentiment. Developers may additionally reimplement some of the modules and reuse the others. Each different setting consists of a new instantiation of PoeTryMe. So far, PoeTryMe has been instantiated to produce poetry in different forms, including song lyrics [7], and in different languages, originally Portuguese [6], and later also Spanish and English [11], given different stimuli, such as Twitter trends [9] or concept maps extracted from text [13].

A limited version of PoeTryMe can currently be used through the TryMe web interface[4] [8], which generates poems given a language, one of the predefined forms, a list of seeds and a surprise factor. Poems are generated following a generate-and-test strategy, also used in most instantiations of PoeTryMe: lines are generated, one after another, until the metre and rhyme constraints are met or a certain number of generations is reached. The generation strategy cannot be changed through TryMe, nor can additional parameters, such as the semantic network, the generation grammar, turning the seed expansion on, or the desired polarity. Yet, this interface enabled the identification of most limitations discussed in the following paragraphs, together with the *Poeta Artificial*[5] Twitter bot [9], which continuously generates Portuguese poems inspired by current trends. We should nevertheless note that some limitations can be minimised if PoeTryMe is used with its full capabilities. For instance, increasing the number of generations, not possible through the interface, increases the chance of rhymes in a trade-off of longer generation time.

## 4 Identified Limitations of PoeTryMe

The main limitation of PoeTryMe is that the meaning and intention of the poems is not clearly-defined. Given seed words constrain the semantic network, so that each line uses them or semantically-related words. Each relation can be rendered as text following a set of patterns in the generation grammar, which apply for every relation of the same kind. Poems will thus be richer for larger semantic networks, not only in terms of words and relation instances, but with a varied range of relation types. The generation grammar also plays an important role, because it should contain several different renderings for each relation type, and for lines with different number of syllables.

---

[4]TryMe section in `http://poetryme.dei.uc.pt`
[5]Check `https://twitter.com/poetartificial`

Although the aforementioned approach enables the generation of semantically-coherent lines, in the strategies implemented so far, they are generated independently. Therefore, although the connection to the seeds provides some consistence and the poem is indeed on a certain topic (check a previous evaluation [11]), the sequence of ideas is not always the best. Also, the same word might have different meanings, which, in some cases, might be interesting, but may also shift the poem semantics. One the other hand, although this feature is not available with the TryMe interface, the system can explain the semantic connection between the selected words and their connection to the seeds, which might be helpful to understand the underlying meaning. This limitation can be further minimised if the semantic network is replaced by one specifically-tailored for the target domain, as in previous work [13], where it was extracted from a given Wikipedia article, but the generation grammar would also have to be tuned, regarding the extracted relations and, especially, the user intention.

The following poem, generated with the seeds *poetry* and *interaction*, illustrates the possible results of PoeTryMe.

> *and interaction aggressions*
> *no more poetry expressions*
> *tag that poetry too loud*
> *action says it, stage says it*

Each line has seven syllables and two of them rhyme, but their sequence is not the most logical. Semantic consistence is aided by the relations rendered in each line, which involve the seeds or related words, namely: coHyponymOf(*interaction*, *aggression*), coHyponymOf(*expression*, *poetry*), domainOf(*tag*, *poetry*), hyponymOf(*action*, *stage*). The last instance is indirectly connected to *interaction*, through hypernymOf(*interaction*, *action*), which is possible because the poem was generated with a surprise factor greater than 0.

This leads us to another limitation of PoeTryMe: it might be tricky to select the surprise factor. This parameter sets the probability of selecting words that, in the semantic network, are more than one level further to the seeds. More precisely, a high surprise factor increases variation in vocabulary by considering more words that are not directly connected with the seeds. On the one hand, a low surprise might result in a poem where every line uses one seed, thus not so interesting. See, for instance, the following example, generated with the seeds *problem* and *solving*, with surprise set to 0.

> *fall like a dead riddle problem*
> *we got solving and locations*
> *solving write determinations*
> *with problem and pitfall on?*

On the other hand, a higher surprise might result in a poem where the connection to the domain is harder to find, such as the following, generated with the seeds *spatial* and *cognition* and a surprise of 0.1 – meaning that every relation instance connected to a word that is connected to a seed has a 10% probability of being used by the Line Generator.

> *if you locate what you place*
> *the mind doubters of her embrace*
> *ears were nasty, heads were hot*
> *chair, moon, and places forgot*

Despite an indirect connection with the seed words, none of them ends up being used. As a probability, the impact of this factor is highly influenced by the number of relation instances involving the seeds and the words directly related to them – e.g. 10% of all the words related to *person* or *place* can be a much higher number than 10% of all the words related to *solving* or *computational*, because the former are nodes with a higher degree in the network. In fact, when the number of relations involving each seed is not balanced, the resulting poem might use only words related to the seed that is involved in more relation instances. This happens, for instance, in the following poem, generated with *computational* and *creativity*, where every line uses the latter or related words (namely, coHyponymOf(*hands*, *creativity*)) and the former is just forgotten:

> *design leads to creativity*
> *creativity plane aptitudes*
> *while the stormy hands do right*
> *creativity spin, flight*

Besides making the poems potentially more interesting, this issue can be minimised if the initial seeds are automatically expanded with a small set of relevant words, obtained with the PageRank algorithm (as in previous work [7, 11]). Yet, again, this feature is not available in the TryMe interface.

A final limitation is related to the grammar renderings. One assumption behind PoeTryMe is that all relation instances of the same type can be expressed by the same textual pattern, changing only the involved words. For instance, "*the <x> of the <y>*" should hold for every pair of words, such that $partOf(x, y)$. However, in order to avoid time-consuming work for handcrafting this kind of grammars, they ended up being automatically extracted from human-created poems and song lyrics, which enabled the creation of larger and richer grammars, but also harder to control. Therefore, poems occasionally present grammatical inconsistencies, odd syntactic constructions, or words from a different semantic domain, that are a fixed part of the pattern. See, for instance, the following poem, generated with the seeds *human*, *creative*, and *cognition*:

> *a imaginative creative sea*
> *a insertion where contents are free*
> *license plate with mankind human on it*
> *filled with joy... human becomes mitt*

In the first two lines, the determiner *a* should instead be *an*, because it precedes words starting with a vowel. Furthermore, words such as *sea* or *license plate* are respectively part of the pattern used and, although might accidentally result in something interesting, do not have a strong connection with the seeds. Another issue is the odd construction *mankind human*.

## 5 Extending PoeTryMe's API

Since TryMe became available, it has been tested by different people and, together with the outputs of the *Poeta Artificial* Twitter bot [9], the previously discussed limitations were either identified or confirmed. In this process, some users pointed out items that would improve the system, often expressing their will to change the resulting poem. For instance, in some cases: they preferred a different line order; some lines did not end in rhyme, or did not match the target metre exactly; punctuation was missing or there was a grammatical inconsistence; the syntax or the semantics was bizarre; they were looking for lines with a deeper meaning; they just wanted to change a word or a line for no specific reason. Some users even confessed to have generated several poems, possibly with different parameters, in order to obtain more suitable lines or words, which they would then manually organise as a new poem. This was our main motivation for providing an alternative way of using PoeTryMe.

Before this work, PoeTryMe's API already had a REST API with a single endpoint:

- `http://poetryme.dei.uc.pt:8080/PoetrymeWeb/rest/poetry`

Used by TryMe and by ConCreTeFlows [22], this endpoint enabled the generation of a full poem from a small set of parameters, namely:

- Language: `lang=[en|pt|es]`

- Form: `form=[id of the form]` (either from a pre-defined list, also including song lyrics; or a string $n \times m$, where $n$ is the number of lines and $m$ is the number of syllables)

- Seeds: `words=[comma-separated list of words]`

- Surprise: `surp=[0-1]`

For instance, the following URL returns a sonnet in Portuguese, using the seeds *criatividade* and *computador*, with a surprise of 0.01:

> `http://poetryme.dei.uc.pt:8080/PoetrymeWeb/rest/poetry?lang=pt&form=sonnet&seeds=computador+`
> `criatividade&surp=0.01`

Towards finer-grained interaction with PoeTryMe, the API was extended with additional endpoints for performing lower-level functionalities, namely:

- Generation of a single line, given a list of seeds and a surprise factor, selected from the best $n$ lines generated, according to a target number of syllables:

  `http://poetryme.dei.uc.pt:8080/PoetrymeWeb/rest/poetry/line`

  - Language: `lang=[en|pt|es]`
  - Seeds: `seeds=[comma-separated list of words]`
  - Surprise: `surp=[0-1]`
  - Target number of syllables: `nsyl=[1-n]`
  - Maximum generations: `bestof=[1-n]`

- Retrieve a set of words related to a target word. Relation can be semantic, same number of syllables, same rhyme, or a combination of the previous. Resulting words might be further constrained with a target polarity:

  `http://poetryme.dei.uc.pt:8080/PoetrymeWeb/rest/poetry/words`

  - Language: `lang=[en|pt|es]`
  - Word: `word=[word]`
  - Relation: `rel=[id for relation type]` (supported types are: `synon` for synonymy, `anton` for antonymy, `hyper` for hypernym, `hypon` for hyponymy, `cohyp` for co-hyponymy, `other` for any other type, or `any` for any type)
  - Target rhyme: `rhyme=[word with the target rhyme]`
  - Target number of syllables: `syl=[word with the target number of syllables]`
  - Polarity: `pol=[-1,0,1]`

- Score a piece of text for a target poetry form. This will use the same metric as PoeTryMe, which means that poems that perfectly match the metre will be score with 0, with (negative) bonus for each pair of rhyming lines, and a (positive) penalty for each syllable out-of-metre:

  `http://poetryme.dei.uc.pt:8080/PoetrymeWeb/rest/poetry/score`

  - Language: `lang=[en|pt|es]`
  - Form: `form=[id of the form]` (same possibilities as for the poem generation URL)
  - Text: `text=[full text]` (lines split with a `\n`)

The new functionalities can be exploited by different systems, which may use them for different approaches to poetry generation or alternative purposes. In the next section, a use case of this API is presented: a co-creative web interface for PoeTryMe that uses these functionalities on demand.

## 6 Co-PoeTryMe: a Co-Creative Interface for PoeTryMe

Co-PoeTryMe [12] is a web-based tool, mostly developed in JavaScript, that enables the human user to interact with PoeTryMe, hopefully towards better poems, or more in line with their intentions. Interaction starts with a draft poem, which may be fully generated by the system, written by the user from scratch, or imported from an existing text file. After this, the user may manually edit parts of the poem, select words or full lines.

Co-PoeTryMe has several visual modules, but they are not all visible at the same time. Figure 1 depicts the modules for poem Edition (centre), Draft generation (left) and Instructions (right) in the current prototype. Once the draft has content, the Edition module becomes visible. The user can double-click to edit the full poem, a line, or a word manually, and can also drag-and-drop words or lines to switch them. The Drafts module has buttons for enabling the generation of a new draft that will replace the current one (play-like button), and to score the current draft based on the metrics and the presence of rhyme, according to the selected poetry form (star-like button). The top module, always visible, has utility buttons for selecting the application language (English or Portuguese), showing or hiding tooltips, importing or exporting a draft, sharing in social networks, undo, redo, as well as a tool for visualising the changes made from the initial draft to its current state.

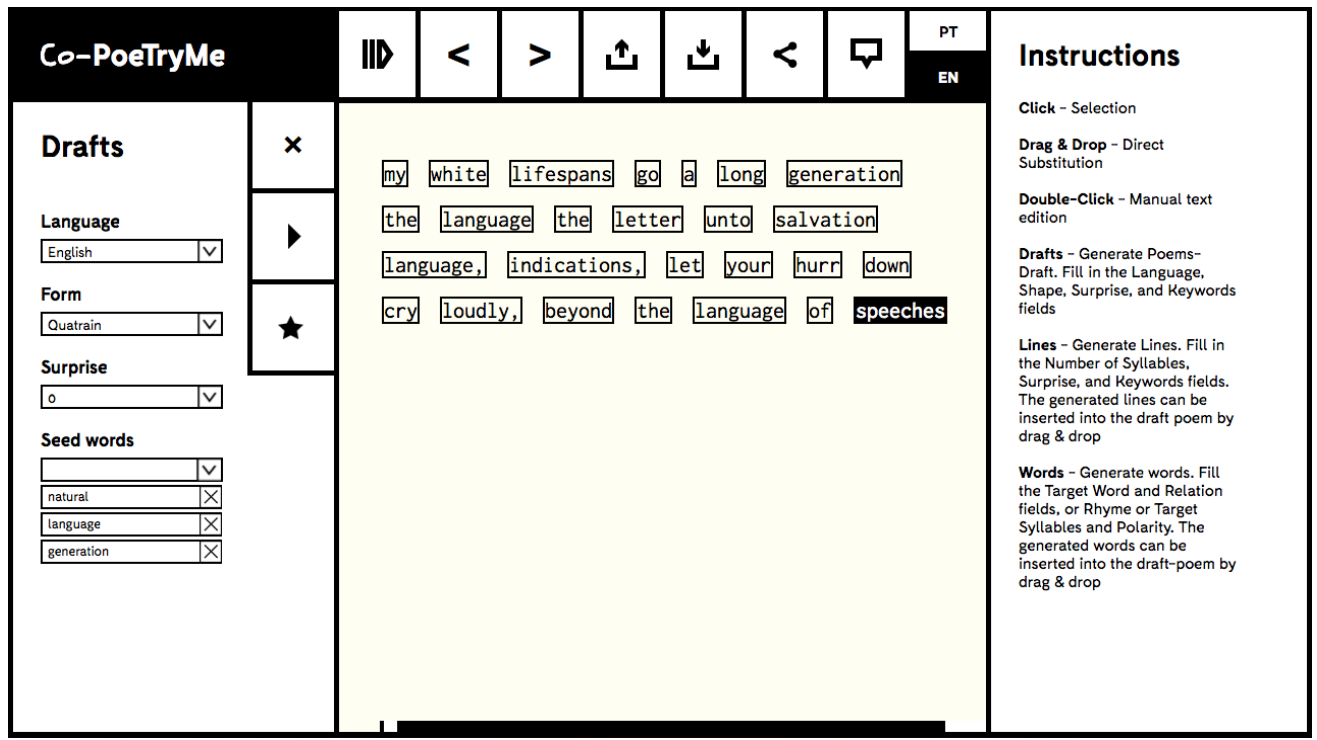

Figure 1: Co-PoeTryMe's prototype: poem edition and word generation.

When a word is clicked, word-related tools are shown instead of the Drafts and Instructions modules, as depicted in figure 2. On the left, the Words module enables the retrieval of words with certain features, such as a semantic relation, rhyme or the same number of syllables as a selected word, possibly with a given polarity. Retrieved words appear in a cloud, located in the Generated Words module, on the right-hand side. They can be drag-and-dropped to be added to the poem, they can replace words in the poem, or they can be moved to the Word Bank module, for future utilisation. Replaced words are moved to the Trash module. At each word generation action, triggered by the play-like button, the words in the Generated module are lost, while words in the Bank and in the Trash remain available for future inclusion in the poem. In figure 2, words related to *language* and rhyming with *salvation* were retrieved.

When a complete line is clicked, the Lines module is shown instead of the Words or the Drafts module. The Lines module enables the generation of new lines and, similarly to the words, it also enables a Generated, a Bank and a Trash module, specific for lines, on the right-hand side. Figure 3 shows the line-related modules while, in the center, the changes made by the user in the original draft are illustrated. Currently, the represented changes cover words and lines switched and words added, removed or replaced. Symbols inspired by document revision tasks are applied, and the text produced by the human has a distinct typography than text produced automatically. Besides a static visual representation of the changes, changes can be animated from the beginning to the current state of the draft, through a play-like button that appears in the bottom of the Edition module. The static representation may also be exported as an image and drafts may be directly shared in social networks.

## 7 Conclusion and Further Work

This paper described how the suggestions of several users of PoeTryMe, a poetry generation system, were followed to enable a finer-grained and more user-driven interaction with this system, while also minimising some of its identified limitations. The modular architecture of PoeTryMe eased the extension of its public API with additional endpoints for performing sub-tasks of poetry generation, such as the generation of single lines or the retrieval of related words. This enables the exploitation of specific modules of PoeTryMe by external systems such as Co-PoeTryMe, a co-creative tool that enables the user to collaborate with PoeTryMe in the composition of new poems.

In the future, we are planning to make more functionalities available through the API, such as the contextualization of lines according to the underlying semantic network. We are still working on the development of Co-PoeTryMe, especially on the usability and design aspects. Usability tests are currently being conducted and will hopefully provide us useful feedback on aspects to improve and directions to take. New functionalities may also be added in a near future (e.g., score the metre of poems, according to a given form). Co-PoeTryMe is accessible online, from a link in PoeTryMe's website, at: `http://poetryme.dei.uc.pt/`.

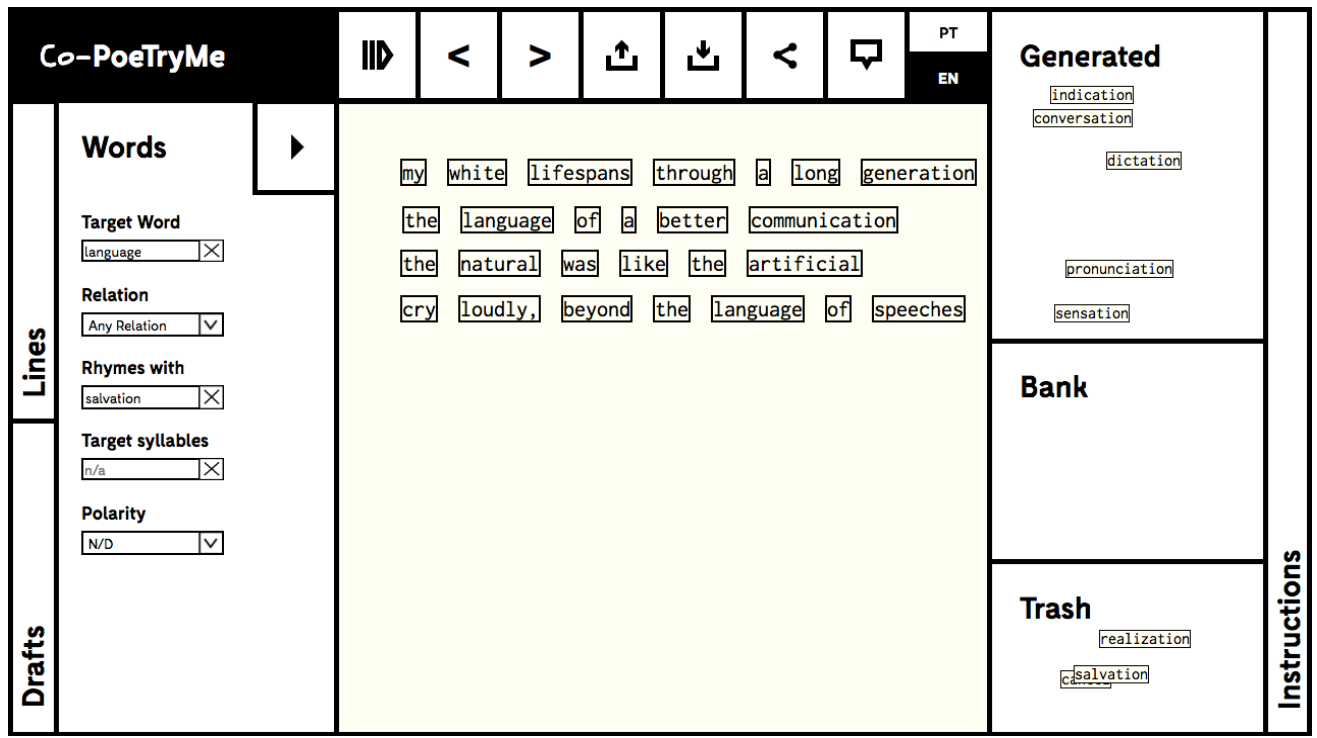

Figure 2: Co-PoeTryMe's prototype: poem edition and word generation.

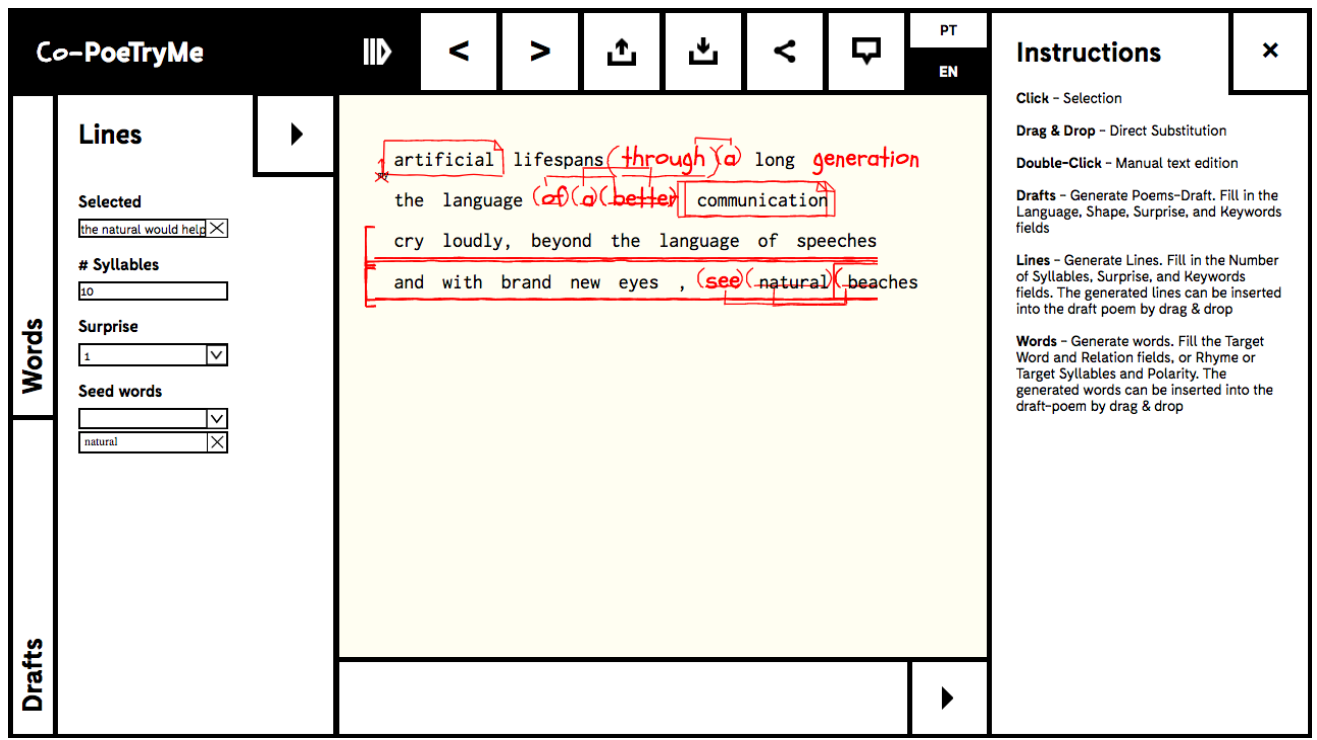

Figure 3: Co-PoeTryMe's prototype: visualising changes.

# Towards the Recognition of Sketches by Learning Common Qualitative Representations

Wael Zakaria[1]    Ahmed M. H. Abdelfattah[1]    Nashwa Abdelghaffar[1]    Nermeen Elsaadany[1]
Nohayr Abdelmoneim[1]    Haythem Ismail[2]    Kai-Uwe Kühnberger[3]

Faculty of Science, Ain Shams University, Egypt    Faculty of Engineering, Cairo University, Egypt
Institute of Cognitive Science, University of Osnabrück, Germany

**Abstract.**

The aim of this paper is to take advantage of major perspectives for processing hand-drawn sketches to build a hybrid technique, in which aspects of machine learning, computer vision, and qualitative representations are mixed to produce a classifier for sketches of four specific types of objects.
**Keywords:** Feature learning - HOG - spatial relations - sketch recognition

## 1   Introduction and Problem Definition

Sketch recognition is a well-established field in AI and spatial cognition, where techniques that recognize hand-drawn sketches usually follow one of two major perspectives for processing common sketches: raster and vector. The former considers a sketch as an image, from which a feature vector is built from a histograms of oriented gradient (HOG) of pixels. Then, it uses a machine learning technique, such as support vector machines, to build a classifier model. Vector processing, on the other hand, deals with a sketch as a set of points, from which qualitative representations are extracted to build a knowledge base of features, on which analogical processing models can be applied for the recognition process. (cf. §4 for some references to existing work in sketch recognition.)

This paper presents a hybrid technique, QuRSeR-Tech, that is based on qualitative representation for recognizing hand-drawn sketches. The presented technique goes somewhat down into low-level representations of sketches, where a sketch consists of geometric shapes as the sketch's constituents that have simple QR features such as relative sizes of geometric shapes, positional relations of any successive pairs of geometric shapes, and similarity group of any number of successive geometric shapes that have the same relative sizes. Our technique has two SVM-based classifiers that employ histograms of oriented gradient, HOG, characteristics from computer vision, and a restricted set of qualitative representations (QR) [Scholkopf and Smola, 2001,Dalal et al., 2006,Cohn and Renz, 2008].

### 1.1   Abstract Building Blocks

In the following, we formulate the basic definitions of the terms we use in this paper, and explain some conventions that help understanding the rest of the presentation and the obtained results.

A stroke is the basic building block of a sketch. It can be represented as a sequence of points in the sense of a polyline, but we formally define a **stroke** as a sequence $s := \langle p_i \rangle_{i=1}^{n(s)}$ of $n(s) \in \mathbb{N}$ ordered tuples that correspond to sequentially recorded points between a pen-down and a pen-up events. Each representation of a point $p_i$ corresponds to a tuple containing both the $x_i$- and $y_i$- coordinates of the pixels composing $s$, as well as a temporal parameter, $t_i$, that reflects the time of drawing each pixel $p_i$, for $1 \leq i \leq n(s)$. For simplicity, we may assume that a stroke $s$ is a sequence of $n(s)$ ordered pairs, $s := \langle (x_i, y_i) \rangle_{i=1}^{n(s)}$, with $p_1 = (x_1, y_1)$ being the pair corresponding to the coordinates of the "start point" of drawing $s$ (pen down), and $p_{n(s)} = (x_{n(s)}, y_{n(s)})$ corresponding to $s$'s last point (pen up). Even though, $p_1$ and $p_{n(s)}$, in particular, must also reflect (at least) the temporal parameters, $t_1$ and $t_{n(s)}$, that record the start and end times, respectively, of $s$. We also define a **sketch** as a sequence of strokes, $\mathcal{S} = \langle s_1, s_2, \ldots, s_{m(\mathcal{S})} \rangle$, where $i$ is the stroke number for each $1 \leq i \leq m(\mathcal{S})$. Two strokes $s_i$ and $s_{i+1}$ are called *successive*. Figure 1a gives an example of a sketch, $\mathcal{S}_{w_1} = \langle s_1, s_2, s_3, s_4, s_5, s_6, s_7 \rangle$ that contains seven strokes (i.e., $m(\mathcal{S}_{w_1}) = 7$) corresponding to two successive "ellipses" (circles) followed by five consecutive "line segments".

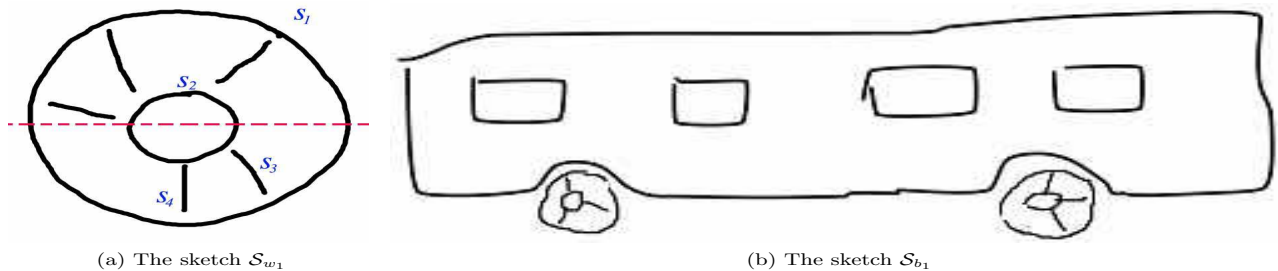

(a) The sketch $\mathcal{S}_{w_1}$    (b) The sketch $\mathcal{S}_{b_1}$

Fig. 1: Two hand-drawn sketches of two different objects: (a) a WHEEL object (which is referred to in the text as $\mathcal{S}_{w_1}$, with some illustrative labels and a dashed line), and (b) a BUS object (referred to as $\mathcal{S}_{b_1}$).

To assist the learner presented later in §2, we base the building on a **feature space** that contains guiding information about all sketches of the set of objects to be learnt (cf. Figure 2). From all the sketched objects, we gather and record a specific number of qualitative representational (QR) criteria (which are limited in this paper to the following three kinds): *relative size*, *positional relation*, and *similarity group*. Relative sizes measure the drawn stokes' metric lengths compared to that of the whole sketch they compose. The size of the (whole) sketch is measured by calculating the perimeter of the canvas that contains it. Based on this size, values ranging from zero to the perimeter's are discretized into intervals, each of which is assigned to one of the **relative sizes**: tiny, small, medium, semi-medium, large, very large, and huge. The relative size of a stroke is assigned to the intervals it belongs to. A **positional relation** is a binary relation between (the centroids of) the two geometric shapes corresponding to successive strokes in a sketch. For a sketch $\mathcal{S} = \langle s_1, s_2, \ldots, s_{m(\mathcal{S})} \rangle$, positional relations $R(s_{i+1}, s_i)$ on $\mathcal{S} \times \mathcal{S}$ determine the location of a stroke $s_{i+1}$, relative to that of the stroke immediately sketched before it, $s_i$, for $1 \leq i < m(\mathcal{S})$, where the positional relation $R$ can be one of the following: equal ($\equiv$), up ($\uparrow$), right ($\rightarrow$), left ($\leftarrow$), up-left ($\nwarrow$), up-right ($\nearrow$), down ($\downarrow$), down-left ($\swarrow$), and down-right ($\searrow$). A **similarity group** is a collection of two or more geometric shapes of the same relative sizes. In other words, successive geometric shapes of the same type (e.g., ellipses) can be grouped together into one group if they have the same relative size.

For each sketch $\mathcal{S} = \langle s_1, s_2, \ldots, s_{m(\mathcal{S})} \rangle$, we record a (constant) number $M$ of selected features that reflect the qualitative representation of a sketch $\mathbf{QR}(\mathcal{S})$. A QR feature vector, $f_\mathcal{S}$, is used for recording numerical values $\langle f_i \rangle_{i=1}^{M}$ as representatives of the stored feature values. The features we use here have the fixed ordering $f_1, f_2, \ldots, f_M$ (cf. §2). If no stroke satisfies any given feature $i$, then $f_i := 0$. The feature vectors of all the $N$ (learning sample) sketches are collected in the feature space $F$. If $f_i \neq 0$, then the selected feature number $i$ is satisfied by exactly $f_i$ strokes (of each of the $N$ sketches). The value $f_i \neq 0$ means that either (i) $f_i$ represents the number of geometric shapes (for the corresponding relative size features), (ii) $f_i$ represents the number of pairs of geometric shapes (for positional relation features), or (iii) $f_i$ represents the number of $n$ geometric shapes (for similarity group $n \geq 2$). Thus, the QR feature space $F$ is a matrix of $N \times (M+1)$ integers, in which the $N$ rows represent the $N$ sketches, while its first $M$ columns represent the $M$ QR feature vectors of the sketches, and the $(M+1)^{\text{st}}$ column is a class value (e.g., the sketched object's label; cf. the "class value" column in the right part of Figure 2).

### 1.2 Example: The $\mathcal{S}_{w_1}$ wheel

The wheel in Figure 1a has been sketched in the following order: outer circle/ellipse, inner ellipse, four lines started at down right and move clockwise. For this object, our technique extracts several features, such that the relative size of the outer ellipse is medium, the smaller ellipse is tiny, and the four lines are tiny. The positional relations between the outer circle and the inner circle is that the inner is equal to the outer (w.r.t. centroid positions), but the first line is down right of the inner ellipse, the second line is down left the first one, etc. The four lines are grouped together and create one group (cf. §1.1). Let $\mathcal{S}_{w_1} = \langle s_i \rangle_{i=1}^{7}$ represent the wheel sketch in Figure 1a. Let $s_1$ and $s_2$ be the outer and inner "circles", respectively, and assume that their centroids have the same location. Let $s_3$ and $s_4$ be the right and left strokes, respectively, below the dashed line in the same figure (that is, $s_4$ is the line stroke immediately below the inner circle $s_2$, and $s_3$ is the stroke to its right in the figure[1]). Instead of the postfix notation, $\equiv (s_2, s_1)$, of the positional relation "equal", we use the infix notation $s_2 \equiv s_1$ to indicate that $s_2$ is located at the same location of $s_1$. Since $s_3$ is a successor to $s_2$, and spatially located to the latter's "down-right", we write $s_2 \searrow s_3$. Similarly, $s_4 \swarrow s_3$ indicates that the semi-vertical line segment, $s_4$, comes "down-left" of (and successor to) the leaned line segment, $s_3$. The five consecutive line strokes $\langle s_i \rangle_{i=3}^{7}$ of the wheel $\mathcal{S}_{w_1}$ of Figure 1a are successive and have the same relative size (which is "tiny" in this case). Therefore, these strokes are combined, forming a group labelled "*group of three or more lines*". The feature vector for $\mathcal{S}_{w_1}$ contains values that reflect the following features of the wheel in Figure 1a: five tiny lines, one tiny circle, and one small circle. This means that the value of feature "tiny-line" is 4, of "tiny-circle" is 1, and of "small-circle" is 1. The similarity group labelled "*group of two circles*" has the value 1.

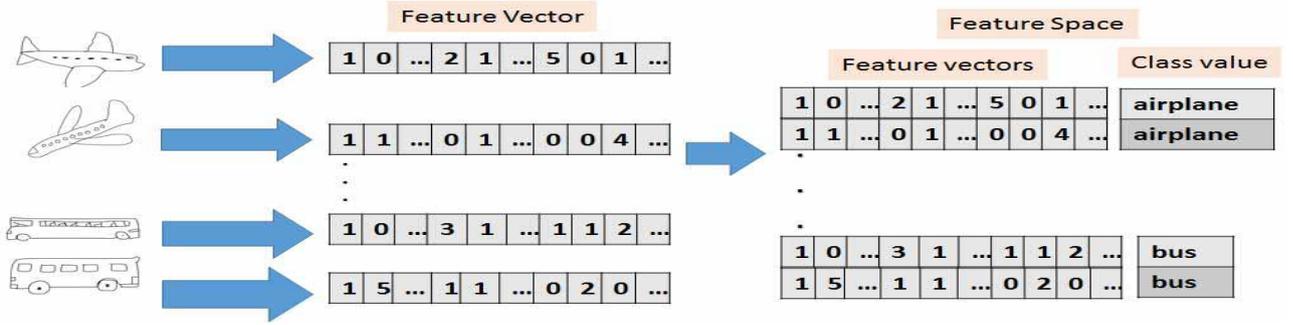

Fig. 2: Constructing the Features Space $F$ from the $N$ sketches.

## 2 Methodology

We present a Qualitative Representation used for Sketch Recognition Technique, or *QuRSeR-Tech* for short, which is proposed to build its classification model by learning a restricted set of common qualitative representations. QuRSeR-Tech is used for recognizing newly drawn sketches based on a previously given set of learning sketches. QuRSeR-Tech consists of two learning systems and, consequently, two kinds of datasets used for learning purposes.

The first classifier utilizes HOG-based extracted features for recognizing drawn strokes within **geometric shapes**. The geometric shapes are considered the low-level representation constituents of a sketch. Here, these geometric shapes are "line", "arc", "ellipse", "poly-line", "triangle", "rectangle", or "polygon". For each geometric shape, we collect a set of hand-drawn sketches describing it. These shapes are used as a training dataset in order to learn and build a classifier model that, in turn, is able to classify a given stroke as one of the geometric shapes. To that aim, we extract the features that discriminate each kind of image. The first dataset contains hand-drawn strokes (the learning samples), each of which is classified into one of the geometric shapes. The second classifier uses QR-based extracted features for recognizing a hand-drawn sketch to an object name. The second dataset contains hand-drawn sketches, each of which is classified into one of the **object names**: "AIRPLANE", "BICYCLE", "BUS", or "HOUSE".

QuRSeR-Tech's first learning system is called *StrokeToGeometric-classifier* (StoG), which builds its classifier model to recognize a hand-drawn stroke (as one of the aforementioned geometric shapes). After that, all recognized geometric shapes and their qualitative representations are stored together, to build a new dataset that will be set as input for the second learning system, which we call *SketchToObject-classifier* (StOb). The latter is a backbone of QuRSeR-Tech, and is is used for classifying a hand-drawn sketch.

---

[1] Note that the labeling of the strokes is intended to reflect the successiveness of the strokes; e.g., $s_4$ is sketched after $s_3$. This is vital, especially because the presented positional relations are not all commutative.

## 2.1 StoG classifier: The $1^{st}$ Learner

The Histogram of Gradient Orientation (HOG) is a method that is based on evaluating well-normalized local histograms of image gradient orientations in a dense grid. HOG's basic idea is that local object appearance and shape can often be characterized rather well by the distribution of local intensity gradients or edge directions, even without precise knowledge of the corresponding gradient or edge positions [Dalal et al., 2006].

In order to extract HOG features of an image, one starts with describing the structure of the image from which the HOG features are extracted. Each gray image is represented by a matrix $\texttt{IMG}_{w \times h}$ of integer numbers, $w$ and $h$, varying between 0 and 255, with $\texttt{IMG}(i,j)$ representing the intensity of the image at pixels $1 \leq i \leq w$ and $1 \leq j \leq h$. The image $\texttt{IMG}$ is resized to $\texttt{IMG}'_{128 \times 128}$. The function, *extractHOGFeatures*[2], is applied to extract the HOG features of the resized image $\texttt{IMG}'$, which gives the ability to set the values of its parameters depending on the needs. The output of this function is the extracted HOG features from a gray image $\texttt{IMG}'_{128 \times 128}$, which returns a 1-by-250 feature vector. These features encode local shape information from regions within an image. Therefore, for each image, we extract 250 features stored as an instance in the dataset. As a result, if we have $k$ images, then we extract a feature space as a matrix called $\texttt{HogF}_{k \times 250}$, with one extra column attached that represents the class value (geometric shape) of each image. So the output is $\texttt{HogF}_{k \times 251}$.

A support vector machine (SVM) [Scholkopf and Smola, 2001] is a supervised learning algorithm capable of building a classifier model by learning from training samples to analyze data used for classification. In our case, SVM employs the extracted feature space $\texttt{HogF}$ (of all images as training samples for the learning process), to produce a classifier model, called StrokeToGeometric-classifier (StoG), which is able to classify a drawn stroke as one of the geometric shapes mentioned earlier. In order to use StoG for testing an unknown classified image $\texttt{UIM}_{w \times M}$, we go through the following steps: (i) resize the image into $\texttt{UIM}'_{128 \times 128}$, (ii) extract its HOG features *UIMHOG* which will be a vector $1 \times 250$, (iii) apply StoG to UIMHOG, where the output will be the class value (geometric name) of this image. Figure 3 gives an example of a real case of a hand-drawn sketch (BUS), in which there are 15 strokes that have been drawn. By applying StoG to each one of the drawn strokes, we are able to classify them into: polygon, ellipse, ellipse, ellipse, line, line, line, ellipse, line, line, line, rectangle, rectangle, rectangle, and rectangle.

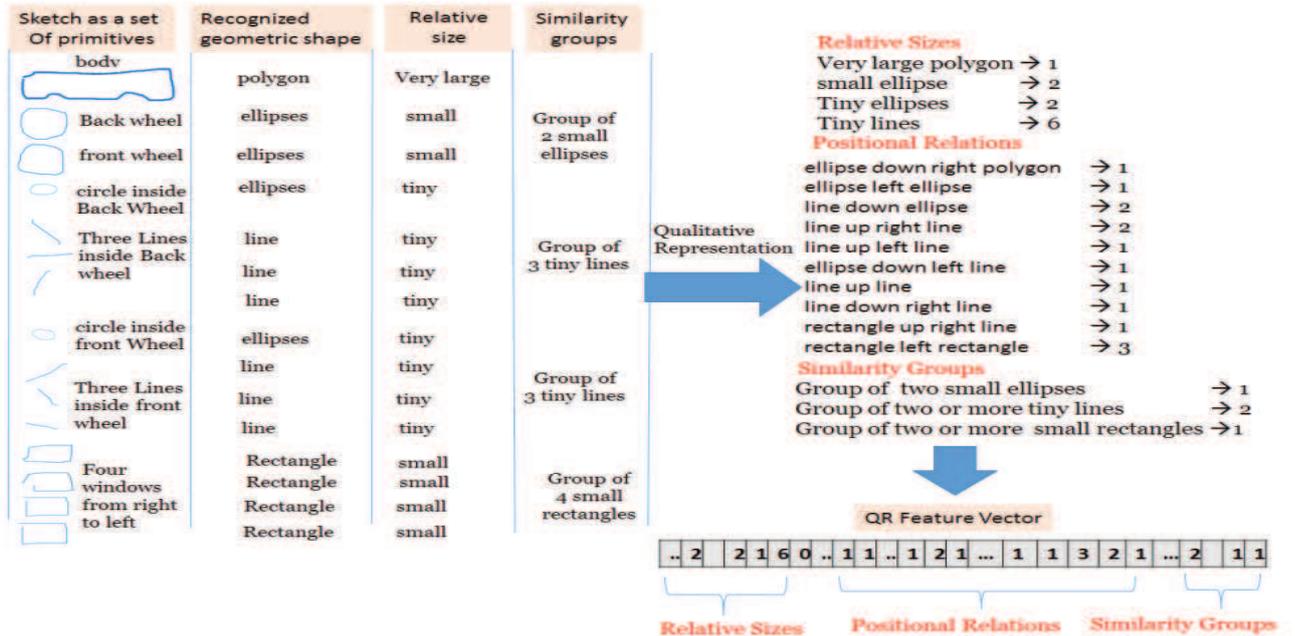

Fig. 3: QR feature vector of the sketched bus, $f_{\mathcal{S}_{b_1}}$, of Figure 1b.

---

[2] This is one of the predefined functions within the Matlab computer vision toolbox.

## 2.2 Qualitative Representation (QR) Identification

Our main goal in this paper is to study the effect of some qualitative representations on building an accurate classifier model that is able to recognize hand-drawn sketches. We present three qualitative representations, along with an argument why they may be considered as suitable candidates. Recall that each sketch $\mathcal{S}$ consists of a set of $m(s)$ strokes, classified into one of the mentioned geometric shapes.

Relative sizes of strokes:

The absolute size of a stroke $s := \langle p_i \rangle_{i=1}^{n(s)}$, called $size(s)$, is the accumulated distances between every two successive points $p_i$ and $p_{i+1}$ belonging to $s$ (for $i < n(s)$). This can be formulated as $size(s) = \sum_{i=1}^{n(s)-1} \sqrt{(x_i - x_{i+1})^2 + (y_i - y_{i+1})^2}$. The relative size is also calculated for each stroke w.r.t. the sketch it belongs to. In our case, we use only relative sizes based on the discretization of the absolute size into one of the following seven discrete values: tiny, small, medium, semi medium, large, very large, huge. We conducted experiments with human subjects, in which the participants draw wheels of busses in different sizes, but almost always approximate a relative size (say "medium") w.r.t. the whole bus sketch. Furthermore, also based on our experiments, rectangles appear (as windows) in the sketches of a house object in large or very large relative sizes, whereas rectangles (as windows) in the sketches of bus objects appear in medium relative sizes (w.r.t. the whole bus sketch). Figure 3 shows that the relative sizes of the drawn strokes: polygon, ellipse, ellipse, ellipse, line, line, line, ellipse, line, line, line, rectangle, rectangle, rectangle, and rectangle are very large, small, small, tiny, tiny, tiny, tiny, tiny, tiny, tiny, tiny, small, small, small, and small, respectively.

We extracted 49 relative size features; for each one of the seven geometric shapes, we use seven relative size features (tiny, small, ..., etc.). Therefore, the relative size features are: tiny lines, ..., huge lines; tiny arcs, ..., huge arcs; ...; tiny polygons, ..., huge polygons. In Figure 3, the relative feature "very large polygon" is 1, which means that there is only one very large drawn polygon, and the relative size feature "tiny lines" is 6, which means that there are 6 drawn tiny lines.

Positional relations:

A positional relation is calculated based on geometric properties of every two successive strokes, $s_i$ and $s_{i+1}$, where the relation is calculated for $s_{i+1}$ with respect to $s_i$ (cf. §1.1). Thus, we make use of 441 features: $s_i \, R \, s_{i+1}$, where $R \in \{\equiv, \rightarrow, \leftarrow, \uparrow, \nwarrow, \nearrow, \downarrow, \swarrow, \searrow\}$ (i.e., $7 \times 9 \times 7 = 441$). An example is shown in Figure 3, where the positional relation called "rectangle $\leftarrow$ rectangle" is set to 3, because there are three pairs of successive strokes having the same positional relation feature.

Similarity groups:

A similarity group is based on the human vision system, and aims to create groups of similar successive strokes. For instance, a participant drew many lines —of similar relative sizes— inside the wheel. These lines are grouped together to form one group, which may contain two geometric shapes or a group of more than two geometric shapes. Hence, we extracted 14 features; each one of the 7 geometric shapes $s$ will have the corresponding 2 kinds of the similarity groups (group of tow geometric shapes or group of three or more geometric shapes). The similarity groups are groups of two lines, groups of three or more lines, ..., group of two polygons, groups of three or more polygons.

## 2.3 StOb classifier: The $2^{nd}$ Learner

The three kinds of extracted features for each sketch are stored together to form a matrix called QR feature space $F$ (as mentioned before in §1). which are 504 (i.e., 49 relative sizes, plus 441 positional relations, and 14 similarity groups).

## 3 Experimental Results and Discussion

We aim at building a classifier model able to recognize a hand-drawn sketch into object name. To achieve this goal, two datasets have been collected through two experiments in order to provide our classification processes with many training datasets from which our learning system can learn and build its model. To evaluate our classifiers, we use a $k$-fold cross-validation technique with $k = 10$.

| StoG classifier | | | | | |
|---|---|---|---|---|---|
| Arc | 91% | | | 4% | 4% |
| ellipse | | 94% | | | 6% |
| line | 10% | | 86% | 5% | |
| polyline | 2% | 2% | 3% | 84% | 10% |
| polygon | 3% | 7% | | 3% | 87% |
| | Arc | ellipse | line | polyline | polygon |

(a)

| StOb classifier | | | | |
|---|---|---|---|---|
| Airplane | 83% | | 10% | 7% |
| Bicycle | 8% | 87% | 5% | |
| Bus | 13% | 5% | 68% | 13% |
| House | 8% | 5% | 5% | 82% |
| | Airplane | Bicycle | Bus | House |

(b)

Fig. 4: (a) Stroke to Geometric shape Classifier's Confusion Matrix. (b) Sketch to Object Classifier's Confusion Matrix.

1. *Hand-drawn Geometric Shapes Dataset:* In the first dataset, we collect many hand-drawn strokes that represent the constituents of any hand-drawn sketch and extract their HOG features to build a HOG feature space. These strokes are classified into one of the following geometric shapes: line, arc, ellipse/circle, poly-line, triangle, rectangle, or polygon. Regarding this issue, a Matlab GUI interface has been implemented for sketching and collecting hand-drawn geometric shapes. Using this interface, the experiment was conducted on 6 subjects where each subject sketched all 5 geometric shapes with different scales, orientations, and positions. We collected 250 hand-drawn geometric shapes and converted them into HOG feature space. The core point is to apply one of the machine learning algorithms such as support vector machine SVM on the HOG feature space, and get a classifier named StoG classifier that will be used for recognizing any sketched stroke and classifying it to only the following geometric shapes line, arc, ellipse, poly-line, polygon. StoG classifier records accuracy= 88.4% for recognizing any hand-drawn geometric shape. Further mathematical analysis are made to identify the polygon to one of the following: triangle, rectangle, or polygon.
Figure 4a shows the StrokeToGeometric-Classifier's confusion matrix, in which the arc and ellipses can be recognized with accuracy greater than 90%, while the other shapes can be recognized with accuracy greater than 83%. We note that 10% of the poly-lines are recognized as polygons, so we use further processes to decide whether the shape is open or closed.[3]

2. *Hand-drawn Sketches Dataset:* In this dataset, we collect many hand-drawn sketches for four specific objects: AIRPLANE, BICYCLE, BUS, and HOUSE. By implementing a matlab GUI interface, a second experiment has been conducted on 12 subjects where each subject drew 20 sketches —five for each one of the four objects. The qualitative representation QR of all those sketches are extracted and stored in QR feature space, $F$. By applying SVM to the feature space, $F$, the sketchToObject-classifier classifies a given hand-drawn object as one the mentioned 4 objects with 80% accuracy. Figure 4b shows the StrokeToGeometric-Classifier's confusion matrix, in which all objects are classified correctly with percentage greater than 80%. In order to show the effects of each QR feature (relative size, similarity group, and positional relation), we build and measure the accuracy of the following 7 models: relative size, similarity group, positional relation, relative size and similarity group, relative size and positional relation, similarity group and positional relation, and all QR. Figure 5 shows that for each one of the 7 cases, 4 classifier model are tested (linear SVM without using PCA for feature selection, linear SVM using PCA, Quadratic SVM without PCA, Quadratic SVM with PCA). The results show that the three combined QR feature plays an important role to achieve the best accuracy. One of the interesting notes is that, without using positional relation, we still have an accurate classifier. This implies that this feature needs more modifications.

All experiments have been implemented and tested on Windows 10, Matlab 2016a, computer vision toolbox, and the classification learner application. All sketches have been collected via WACOM pen, WACOM touch screen, and some sketches from the dataset in [Eitz et al., 2012] are used as guiding images to the participants.

## 4 Conclusion

Understanding what hand-drawn sketches represent, in a way similar to what humans do, is a challenging problem for computation. The representation of sketches using qualitative features plays a key role in this understanding. Qualitative models that can be used to represent sketches are mainly based on two aspects. The first aspect is the way of segmenting an object's contour into meaningful edges. The second aspect is the level of detail for describing an object. The way of handling these aspects can distinguish one model from another.

---
[3] E.g., the distance between the start- and end- points of the drawn stroke is compared to the length of the drawn stroke.

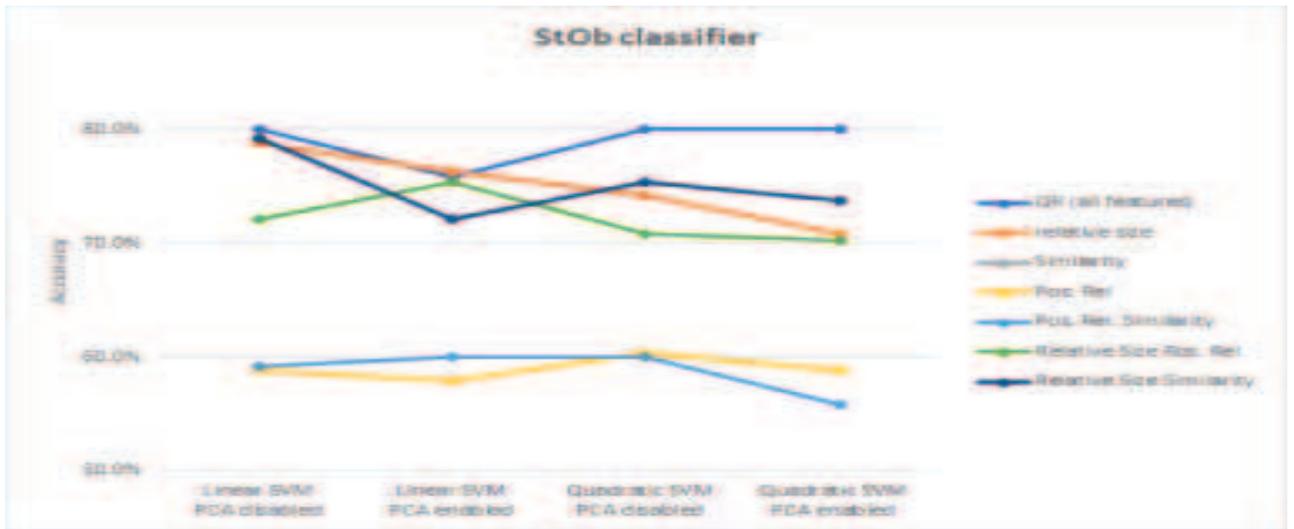

Fig. 5: SketchToObject classifiers accuracy

Here, a technique called *QuRSeR-Tech* has been proposed and implemented for recognizing hand-drawn sketch. Our technique consists of two main stages: StoG classifier (that uses the function extractHOGFeatures for extracting the HOG features of each stroke), and quadratic support vector machine SVM (that is based on the HOG features for recognizing the drawn stroke to geometric shape). To get an accurate classifier, we set the parameters cell size, block size, block overlap, number of bins, and the signed orientation to [4 4], [2 2], [50% 50%], 8, and false respectively. The StoG classifier recorded 88.4% accuracy for recognizing the strokes to geometric shapes. Second, the StOb classifier builds a new representation for any hand drawn sketch by recording three kinds of qualitative representation: relative size, positional relation, and group of similarity. Based on this new representation and quadratic support vector machine, the StOb classifier is able to recognize hand drawn sketches with accuracy 80%. Some comparative studies have been conducted to see the effect each one of the three kinds of QR, in which we proved that the StOb classifier's accuracy will be decreased if any of the mentioned representations is missing. In both of the two classifiers, we apply PCA to reduce the features to only the features that are highly correlated to the class value.

The ideas of this paper are in connection with others in the literature (such as [Wang et al., 2013], [McLure et al., 2011], [Paulson and Hammond, 2008], and [Lovett et al., 2006], to mention a few). [Wang et al., 2013] gives a method based on fuzzy hybrid-based features to classify strokes into geometry primitives. Their human computer interactive system is developed for determining the ambiguous results and then revising the missrecognitions, where their system is developed to classify eight primitive shapes. PaleoSketch is also a primitive shape recognizer that classifies single strokes into certain primitive geometric shapes (cf. [Paulson and Hammond, 2008]). [McLure et al., 2011] describe a higher-level of qualitative representation based on *edge-cycles*, which are sequences of edges connected end-to-end, whose last edge connects back to the first edge. The results of the edge-cycle representation outperforms that of the edge-level representation —produced by CogSketch in [Lovett et al., 2006]— in learning to classify hand-drawn sketches of everyday objects. Although the remarkable performance of edge-cycles representation, it has its own set of problems that can lead to misclassification of sketched concepts (cf. [McLure et al., 2011]).

# Using Stories to Create Qualitative Representations of Motion

Juan Purcalla Arrufi, Alexandra Kirsch

Human-Computer-Interaction and Artificial Intelligence, Universität Tübingen,
Sand 14, 72076 Tübingen, Germany

## Abstract

Qualitative representations of motion transform kinematic floating point data into a finite set of concepts. Their main advantage is that they usually reflect a human understanding of the moving system, so we can more straightforwardly implement human-like navigation rules; in addition, they reduce the overhead of floating point computations. Therefore, they are an asset for mobile robots or unmanned vehicles—both terrestrial and aerial—especially those that interact with humans. In this paper we provide a method to create new qualitative representations of motion from any qualitative spatial representation by using a story-based approach.

## 1 Introduction

Description and interpretation of moving entities (humans, animals, robots, or inert objects) are at the core of many disciplines such as mobile robotics, human-robot interaction, geographic information systems, animal behaviour, high-level computer vision, and knowledge representation, among others. Qualitative representations transform the mass of quantitative data (positions and velocities) into a reduced group of concepts. Therefore, they simplify data so that these are easier to understand and to process (e.g. in modelling, planning, learning, or control).

Nonetheless, the work in qualitative representations of *motion* is still reduced in number, when compared to *spatial* representations[1] [6, p. 16 ][7, p. 5187], and mostly restricted to point-like entities moving in one or two dimensions [22]. Moreover, spatial representations deal with regions [20] and three or more dimensions [11, 1], but this is unusual in representations of motion.

To fill the gap, in this paper we profit from the available spatial representations to systematically increase the number of representations of motion: we introduce a method that creates qualitative representations of motion given any qualitative spatial representation.

This has direct applications, for example, we may create a representation of motion using Hall's spatial categorisation, proxemics [14], which is based on the social distances. Such a representation of motion would describe trajectories according to the personal space and, thus, it could be used to make robot navigation in human environments more friendly.

Our method centres on the concept of 'stories' which, we believe, opens a new perspective in dealing with representations. A spatial representation can classify two static entities, or equivalently, each snapshot of two moving entities. If we therefore consider the complete sequence of snapshots—what we call the 'story' (Def. 2)—, we have a qualitative description of the motion.

---

[1]Wherever we mention the term 'representation' throughout this paper, it is understood that we are talking about '*qualitative* representations' — We drop the term 'qualitative' for sake of readability.

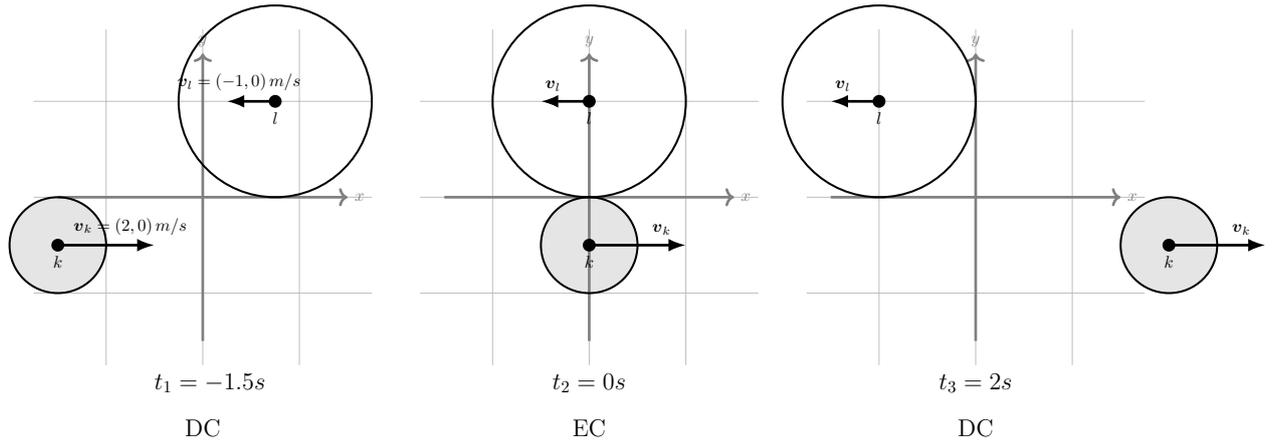

Figure 1: **An RCC story with three qualitative relations.** Two circular entities $k$ (radius = $1m$) and $l$ (radius $2m$) moving with velocities $\vec{v}_k = (2,0)\,m/s$ and $\vec{v}_l = (-1,0)\,m/s$ in the interval $[t_1, t_3]$. The snapshots depict the *temporal sequence of relations* (DC, EC, DC) in the qualitative representation RCC (Fig. 3).

Our method can use any spatial representation (e.g., OPRA$_m$ [17], Rectangle Algebra [3]); however, it can be hampered by the generation of the *stories set* (Def. 3), because this is often an arduous manual task. For that reason we used as a example (Ex. 5) the simple and well-known spatial representation RCC [20] (See Fig. 3). As RCC relates regions, our method will generate, in this particular case, a novel representation of motion—we call it 'Motion-RCC' (Eq. (1) on p. 6)—that deals with regions, and an extended variant, 'Augmented-Motion-RCC' (Eq. (2) on p. 6).

## 2 Related Work

### 2.1 Qualitative Representations of Motion

An overview of representations is found in a survey by Dylla et al. [9]: in a total of 40 representations surveyed they classify three as representations of motion: QRPC [13], RfDL-3-12 [16], and, the most used, QTC [22]. The survey of spatial representations of Chen et al. [4] also mentions three motion representations: Dipole Calculus [18], DIA [21], and QTC.

Representations of orientation and relative direction, such as OPRA [17] or Dipole Calculus [18], are sometimes used to represent moving entities; nevertheless, they are not primarily intended for such a task.

All the aforementioned representations are limited to point-like entities moving in one or two dimensions. There is, however, a particular qualitative relation of motion for regions [23] that is built combining RCC and distances.

### 2.2 Sequences of Qualitative Relations

Continuous sequences of qualitative relations, such as the *temporal* sequences of Def. 1 (p. 4), are based on Freska's foundational concept *conceptual neighbourhood* [12]. Connecting the qualitative relations of a certain representation that are conceptual neighbours we obtain the *conceptual neighbourhood graph*[10] (See example in Fig. 3). So paths in the conceptual neighbourhood graph and continuous sequences of qualitative relations are equivalent.

Sequences of relations are used to analyse real data by Delafontaine et al. [7], and specifically in human-robot interaction by Hanheide et al. [15] from which we borrow the term *'temporal sequence of qualitative relations'* (Def. 1).

## 3 Temporal Sequences of Relations and Stories

In this section, we define and illustrate the key concepts—*stories* and *stories set*—that we use to create qualitative representations of motion (Sect. 5). But first of all we define the underlying concept: *temporal sequence of relations*.

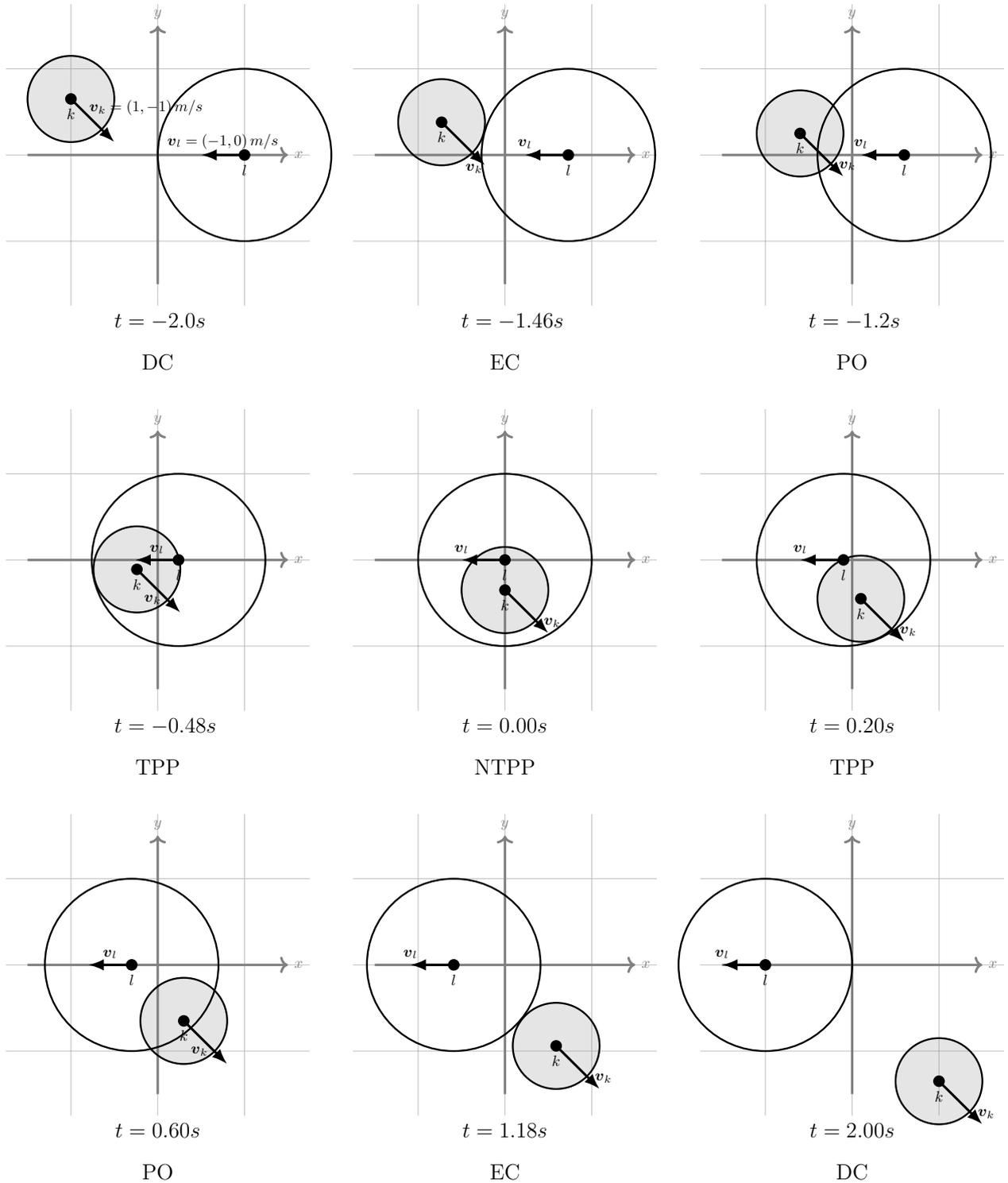

Figure 2: **A RCC story with nine qualitative relations.** Two circular entities $k$ (radius = $1m$) and $l$ (radius $2m$) moving in uniform motion with velocities $\vec{v}_k = (1,-1)\, m/s$ and $\vec{v}_l = (-1,0)\, m/s$. They depict the Temporal Sequence of Relations (DC, EC, PO, TPP, NTPP, TPP, PO, EC, DC) in the qualitative representation RCC (Fig. 3). The snapshots correspond to different increasing times.
This sequence is a *story*, because it remains the same, even if we extend the interval to $(-\infty, +\infty)$. It corresponds to the story $S_5$ of the created representation of motion 'Motion-RCC' (Sect. 5).

**Definition 1.** A **Temporal Sequence of Relations** [15] is a chronologically ordered sequence of qualitative relations of any kind, e.g., space or motion, generated by the motion of two entities in a *time interval* $(t_a, t_b)$.

The time interval $(t_a, t_b)$ can be freely chosen, e.g., it can be totally unbounded, i.e., extend to the whole time $(-\infty, \infty)$, be half-bounded $(-\infty, t_b)$, or bounded $(t_a, t_b)$.

We obtain the temporal sequence of relations of two entities in a certain time interval by mapping their trajectories $\vec{x}_k(t)$ and $\vec{x}_l(t)$ into the qualitative relations of the representation we are using. We describe a sequence of relations as a list in parenthesis: $(R_1, R_2, \ldots, R_i, \ldots)$. We say a temporal sequence of relations is *finite*, if it has a finite number of relations, or *infinite*, if it has an infinite number. Notice that even though the entities' motion occurs in a continuous space throughout a continuous time interval, the temporal sequences are finite, when the trajectories have a finite number of transitions between qualitative relations; this happens in Fig. 1, the sequence is finite, (DC, EC, DC), because we have only two transitions: DC → EC and EC → DC.

Now, based on the temporal sequences, we define the *stories*.

**Definition 2.** A **Story** is a temporal sequence of relations of two entities that is defined over the whole unbounded time interval $(-\infty, \infty)$.

A story describes the qualitative relation of two moving entities at any instant of time. Thus, any temporal sequence of relations is a substring of a certain story. We can see each story as a complete qualitative description of the motion of a two-entities system. We characterise stories with the letter $S$ and, if necessary, an appropriate subscript.

**Example 1.** The temporal sequence $S =$(DC, EC, DC) in Fig. 1 is a story. Any proper substring is not a story, but just a temporal sequence of relations, because it does not happen in the whole unbounded interval $(-\infty, \infty)$. For instance, the substring (EC,DC) is not a story, because it happens on $[0, +\infty)$.

**Example 2.** The temporal sequence (DC, EC, PO, TPP, NTPP, TPP, PO, EC, DC) in Fig. 2 is a story. Substrings, such as (PO, TPP, NTPP, TPP) or (DC, EC, PO), are not stories, but just temporal sequence of relations.

**Definition 3.** The **Stories Set** is the set of all possible stories of two entities.

If there is no constraint on the stories, the stories set contains an infinite number of stories. We refer to the *stories set* with the letter $\Sigma$ (see Sect. 5); we add a subscript, e.g., $\Sigma_0$, when we deal with a set of stories that is not the stories set, but a subset thereof.

## 4 Restricting the Stories: Uniform Motion

The central idea of this paper is to classify motions through *stories*: we assign the same category to the motions that belong to the same story. (Sect. 5). Thus, the total number of categories in our novel motion representation is the cardinality of the *stories set*, i.e., its number of elements. However, an awkward situation arises: the cardinality of the stories set is infinite—some stories are also infinite—, if we do not restrict the motions that create the stories.

Consequently, we suggest restricting the type of motions considered in order to obtain a tractable motion representation. We choose to restrict the stories by considering, from now on, only *uniform motion*, i.e., the velocity vectors are *constant*. This has two desirable properties:

 i. Each story in uniform motion is *finite*, i.e., has a finite number of relations (See Prop. 1 in Appendix A).

 ii. The set of all possible stories in uniform motion, i.e., the *stories set* (Def. 3), is *finite* (See Prop. 2 in in Appendix A). Consequently it partitions the whole *phase space*, i.e., the coordinates space of the positions and velocities of the two entities $(\vec{x}_k, \vec{v}_k; \vec{x}_l, \vec{v}_l)$.

The restriction to uniform motion *stories* is a standard assumption, if we classify motion situations that are specified only by the current position and velocity of two entities, i.e., $(\vec{x}_k, \vec{v}_k; \vec{x}_l, \vec{v}_l)$—the acceleration is disregarded, as in QTC [22]. Though we note that our method may remain valid with other kind of restrictions.

**Definition 4.** A **Rigid Story** is the story of two entities that move with the same velocity, i.e., $\vec{v}_k = \vec{v}_l$.

*Rigid stories* play a special role in uniform motion: each of them is a *singleton*—it has a single element, a constant spatial relation. But not all singleton stories are rigid, e.g., the story $S_{11} =$ (DC) is not rigid but is a singleton (Fig. 4).

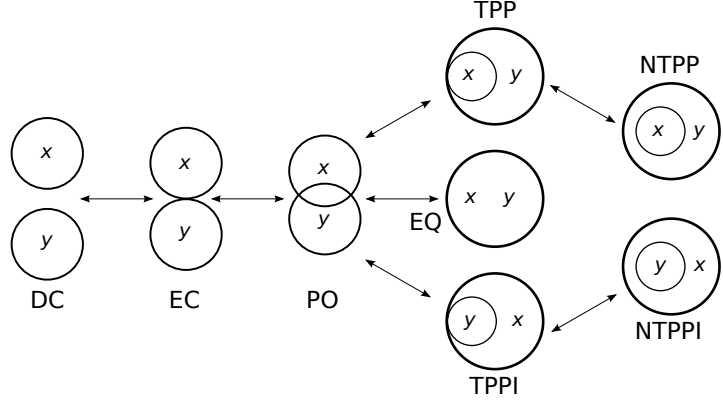

Figure 3: The RCC qualitative relations depend on how two entities overlap. This Figure depicts the 8 RCC relations: DC, EC, PO, TPP, NTPP, EQ, TPPI, and NTPPI as a *conceptual neighbourhood graph* [12, 10]: the arrows connect relations that are *conceptual neighbours* [12] — we switch between conceptual neighbours by a continuous translation without going through any other relation.

## 5 Creating New Qualitative Representations of Motions

We describe the method to create a representation of motion from any given spatial representation. In practice, our method yields always two representations of motion: the simple one, which is just formed by the stories, and the *augmented* variant, which is refined by adding the spatial relations to each story—we combine the power of 'story' and 'snapshots'. We illustrate the method in the example below using the spatial representation RCC (Fig. 3), and thus, the two new generated representations of motion are *Motion-RCC* (Eq. (1) on page 6, and Fig. 4), and its augmented variant *Augmented-Motion-RCC* (Eq. (2) on page 6).

The method is as follows:

1. We have a spatial representation.

2. We calculate the *stories set*, $\Sigma$, for the given spatial representation. In case it is a finite set, e.g., when restricted to uniform motion, we can work out a method to calculate it.

3. The obtained *stories set* is a novel representation of motion, where each story is a qualitative relation—every motion state is classified according to the story it belongs to.

4. (optional) We can create the *augmented representation* of motion from the first one by specifying the spatial relations in each story.

**Example: Creating a representation of motion from RCC**

We illustrate the method above using the spatial representation RCC. (Fig. 3). RCC relates two regions according to their overlapping. So it yields 8 possible relations: DC, regions do not overlap; EC, regions are tangent non-overlapping; PO, regions overlap in the interior but none is contained in the other; TPP, region x is contained in y and is tangent to the border; TPPI, region y is contained in x and is tangent to the border; EQ, both regions overlap completely; NTPP, x is contained in y and does not overlap the border of y; NTPPI, y is contained in x and does not overlap the border of x.

1. We have the spatial representation RCC

2. We calculate the RCC *stories set* restricted to uniform motion as $\Sigma = \Sigma_0 \cup \Sigma_1$. $\Sigma_0 = \{(DC), (EC), (PO), (TPP), (NTPP)\}$ are the *rigid stories* and $\Sigma_1 =\{(DC), (DC, EC, DC), (DC, EC, PO, EC, DC), (DC, EC, PO, TPP, PO, EC, DC), (DC, EC, PO, TPP, NTPP, TPP, PO, EC, DC)\}$ are the non-rigid stories. We rename the rigid stories into $S_{0i}$, $\Sigma_0 = \{S_{01}, S_{02}, S_{03}, S_{04}, S_{05}\}$, and the non-rigid which we rename into $S_{1i}$, $\Sigma_1 = \{S_{11}, S_{12}, S_{13}, S_{14}, S_{15}\}$ according to Fig. 4.

3. The stories set $\Sigma$ is the qualitative representation of motion—note, though, that that story $S_{01}$ and $S_{11}$ are equal to $(DC)$ therefore $S_{01}$ drops to avoid repetition. We call this representation 'Motion-RCC':

$$\textbf{Motion-RCC} = \{S_{02}, S_{03}, S_{04}, S_{05}, S_{11}, S_{12}, S_{13}, S_{14}, S_{15}\} \tag{1}$$

This representation assigns to every motion state $(\vec{x}_k, \vec{v}_k; \vec{x}_l, \vec{v}_l)$ the corresponding story $S_i$, i.e., the corresponding relation of motion.

4. (optional) We can augment the resolution of the representation of motion *Motion-RCC* by specifying the spatial relations in each story—for the singleton stories this process is redundant, as they have a single spatial relation. So we obtain the representation of motion 'Augmented-Motion-RCC'.

$$\begin{aligned}\textbf{Augmented-Motion-RCC} = \{ \\ S_{02}(\text{EC}), S_{03}(\text{PO}), S_{04}(\text{TPP}), S_{05}(\text{NTPP}), \\ S_{11}(\text{DC}), S_{12}(\text{DC}_-), S_{12}(\text{EC}), S_{12}(\text{DC}_+), \\ S_{13}(\text{DC}_-), S_{13}(\text{EC}_-), S_{13}(\text{PO}), S_{13}(\text{EC}_+), S_{13}(\text{DC}_+), \\ S_{14}(\text{DC}_-), S_{14}(\text{EC}_-), S_{14}(\text{PO}_-), S_{14}(\text{TPP}), \\ S_{14}(\text{PO}_+), S_{14}(\text{EC}_+), S_{14}(\text{DC}_+), \\ S_{15}(\text{DC}_-), S_{15}(\text{EC}_-), S_{15}(\text{PO}_-), S_{15}(\text{TPP}_-), S_{15}(\text{NTPP}), \\ S_{15}(\text{TPP}_+), S_{15}(\text{PO}_+), S_{15}(\text{EC}_+), S_{15}(\text{DC}_+)\}\end{aligned} \tag{2}$$

For example, the relation $S_{12}(\text{EC})$ indicates that the entities are moving in the story $S_{12}$ at the moment of tangency, i.e., EC. If the spatial relation appears multiple times in the story, such as EC in $S_3$, we distinguish each appearance, for example, $S_{13}(\text{EC}_-)$ is chronologically the first EC, and $S_{13}(\text{EC}_+)$, the last EC.

## 6 Applications of Qualitative Representations of Motion

We outline two possible applications of qualitative representations

- *Recognition of trajectories (i.e., motion patterns)*
  Through the qualitative relations in the new representation of motion, we can characterise and therefore recognise certain types of motion [7, 15], for example an 'avoidance manoeuvre', as in Eq. (3). This motion sequence begins with the collision story, $S_{15}(\text{DC}_-)$, and ends with a collision free story, $S_{11}(\text{DC})$—the augmented indices, DC, show that nowhere a collision takes place.

$$S_{15}(\text{DC}_-) \to S_{14}(\text{DC}_-) \to S_{13}(\text{DC}_-) \to S_{12}(\text{DC}_-) \to S_{11}(\text{DC}) \tag{3}$$

- *Trajectory control*
  We can use the *conceptual neighbourhood graph* of our new representation of motion to take decisions in order to control trajectories [8]. For example, in the case of Motion-RCC, if we want to avoid a collision we have necessarily to reach the relation $S_{11}(\text{DC})$. Accordingly, the shortest paths in the conceptual neighbourhood graph leading to the relation $S_{11}(\text{DC})$ may provide the needed control operations to avoid the collision.

## 7 Discussion

We have presented a a *story-based* method (Sect. 5) that should be able to generate qualitative representations of motion out of any spatial representation. The created representation of motion inherits the properties of the used spatial representation, e.g., dimensions, or type of entities considered. The method has proven to be effective to generate meaningful qualitative representations of motions for the representation RCC (Sect. 5).

With our generated motion representation, *Augmented-Motion-RCC*, we have outlined two applications of motion representations: recognition of trajectories, i.e., motion patterns; and control of trajectories.

Our generating method is most effective, when we restrict the trajectories of the entities, e.g., setting velocity constant, so that our stories set is finite. This can be seen as a limitation or as the advantage to tailor the generated representation of motion to the features of our trajectories. We have restricted the trajectories to have uniform motion.

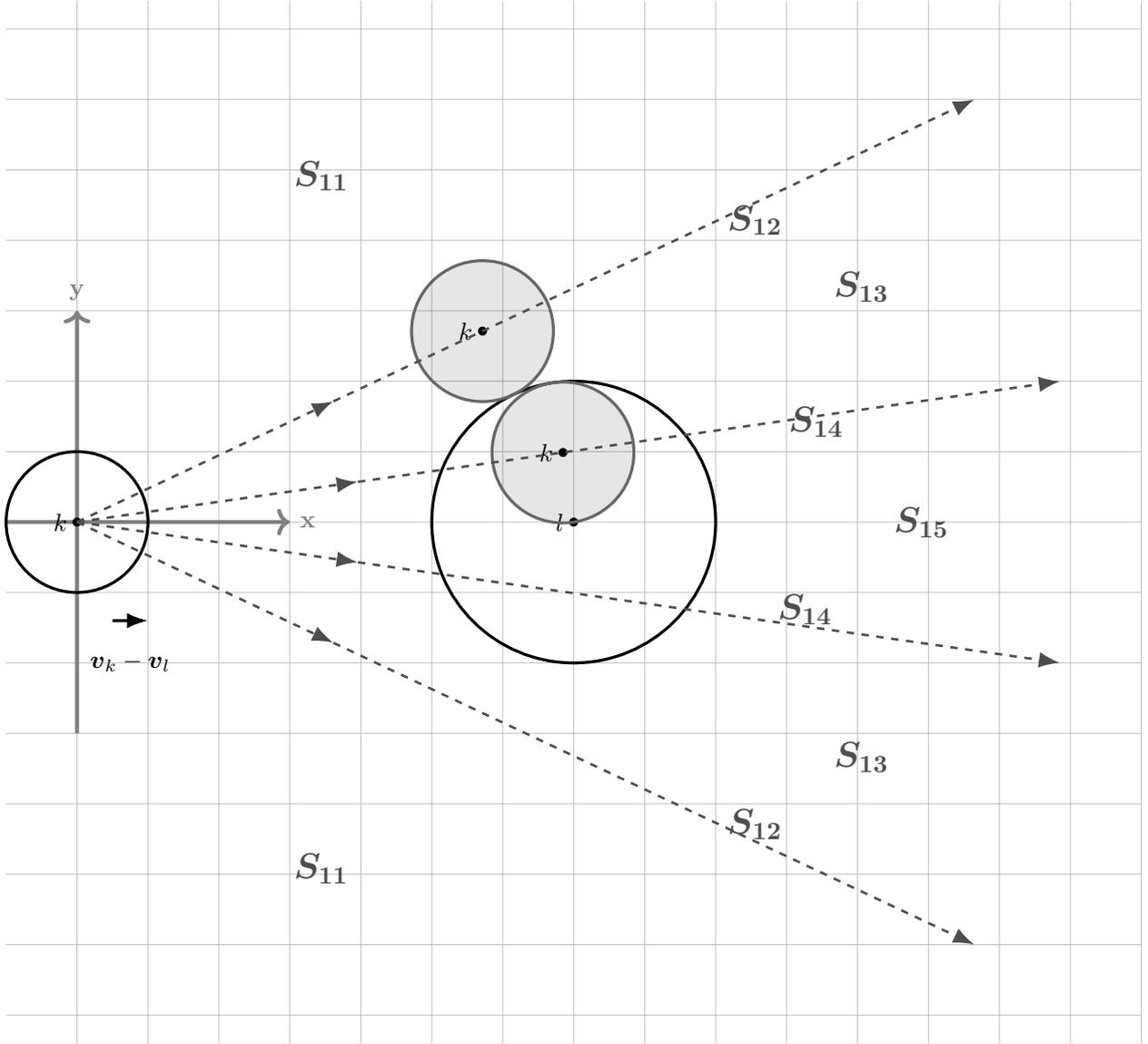

Figure 4: In the representation RCC these are all the possible non-rigid stories, $\Sigma_1$, i.e, the stories of two circles $k$ and $l$ moving in uniform motion with velocities $\vec{v}_k$ and $\vec{v}_l$, so that $\vec{v}_k \neq \vec{v}_l$. The total number is five. Two stories are associated with directions: $S_{12} = $ (DC, EC, DC); $S_{14} = $ (DC, EC, PO, TPP, PO, EC, DC). The remaining three stories are associated with the regions between the directions: $S_{11} = $ (DC), $S_{13} = $ (DC, EC, PO, EC, DC), $S_{15} = $ (DC, EC, PO, TPP, NTPP, TPP, PO, EC, DC).
Note: The figure represents an equivalent simplification that considers $l$ being motionless and $k$ moving with the difference of velocities $\vec{v}_{kl} = \vec{v}_k - \vec{v}_l$. The story depends on the direction of $\vec{v}_{kl}$.

We argue that the use of 'stories' to classify motions borrows from a cognitive idea: we can better recall a series of items, when they are linked by way of a story—Stories seem quite a natural way for humans to relate, connect, or classify items.

The next step are to test the effectiveness of this method with other spatial representations, for instance, three dimensional [11] or those dealing with orientation [19]. Further, we can add the transition time between relations in the stories to obtain quantitative results.

## A  Appendix

**Proposition 1.** *Finitude of the Stories in Uniform Motion*
*We can reasonably show that for two* regular enough[2] *entities the stories in uniform motion are finite.*

We build the proof on two properties: first, stories in uniform motion have *extreme relations* (Lemma 1); second, temporal sequences of relations in uniform motion are finite over a finite time interval (Lemma 2).

*Proof.* According to Lemma 1 two regular enough entities in uniform motion have extreme relations. That is, we can find two time instants $t_a$ and $t_b$, with $t_a < t_b$, so that in the time interval $(-\infty, t_a)$ the entities' relation remains constant—we call it $r_a$—and in the time interval $(t_b, +\infty)$ the entities' relation remains constant.—we call it $r_b$.

Now, According to Lemma 2, these regular enough entities moving in uniform motion have a finite temporal sequence of relations in the interval $[t_a, t_b]$, say $(r_1, \ldots, r_n)$.

Consequently the story of the two entities, i.e., the temporal sequence of relations in the interval $(-\infty, t_a) \bigcup [t_a, t_b] \bigcup (t_b, \infty)$, would be finite, as it is obtained by concatenating the two extreme relations and the temporal sequence: $(r_a, r_1, \ldots, r_n, r_b)$. In case any extreme relation coincides with its border relation, i.e., $r_a = r_1$ or $r_b = r_n$, we exclude the repeated ones. □

**Definition 5. Extreme Relations**
The *extreme relations* are those relations of a story that remain unchanged when $t \to -\infty$ or $t \to +\infty$. That is, a relation $r_a$ is extreme in $t \to -\infty$, if and only if $\exists t_a$, so that in the time interval $(-\infty, t_a)$ the relation between entities is $r_a$. Analogously, a relation $r_b$ is extreme in $t \to +\infty$ if and only if $\exists t_b$, so that in the time interval $(t_b, +\infty)$ the relation between entities is $r_b$.

**Lemma 1.** *Existence of extreme relations for two entities in uniform motion.*
*Two* regular enough[2] *entities that move in uniform motion and are described by a qualitative representation based on overlapping, intersection, or orientation, have a story with extreme relations both for $t \to -\infty$ and $t \to +\infty$.*

*Proof.* We name the entities $k$ and $l$ and they have constant velocities $\vec{v}_k$ and $\vec{v}_l$.

1. In the case $\vec{v}_k = \vec{v}_l$ the relation between both entities, $r_i$, remains constant — this relation is the whole story —, therefore, trivially, $r_i$ is the extreme relation for both $t \to -\infty$ and $t \to +\infty$.

2. In the case $\vec{v}_k \neq \vec{v}_l$ we distinguish two subcases regarding what feature the representation bases on: overlapping-intersection of entities, or relative orientation.

    (a) Representations based on *overlapping-intersection* of finite entities have either one or two qualitative relations for the case of 'no overlapping-intersection', e.g., the relation $DC$ in RCC (Fig. 3); the relation *disjoint* in 9-Int [11]; or the relations '<' and '>' in Allen's Algebra [2]. The mentioned relations must be the extreme relations for each representation, because the distance between two entities that move at different velocities tends to infinity for $t \to \pm\infty$; and consequently the entities do not overlap-intersect any more.

    (b) Representations based on *relative orientation* between entities use the *connecting unit vector* between them, i.e., $\vec{kl}(t) = \frac{\vec{x}_l(t) - \vec{x}_k(t)}{\|\vec{x}_l(t) - \vec{x}_k(t)\|}$, for which in uniform motion, i.e., $\vec{x}_k(t) = \vec{v}_k t + \vec{x}_{k0}$ and $\vec{x}_l(t) = \vec{v}_l t + \vec{x}_{l0}$, we obtain both limits :

$$\lim_{t \to +\infty} \vec{kl}(t) = \frac{\vec{v}_l - \vec{v}_k}{\|\vec{v}_l - \vec{v}_k\|} \ (4a) \qquad \lim_{t \to -\infty} \vec{kl}(t) = -\lim_{t \to +\infty} \vec{kl}(t) \ (4b)$$

---

[2] *Enough regular* entities are those finite in size with a finite number of features, i.e., a finite number of vertices, edges, concavities, holes, ...

Because both limits for the connecting vector exist, the extreme relations of any story exist; they are the relations neighbouring each limit.

□

**Lemma 2.** *Finitude of the Temporal Sequences of Relations in Finite Time Intervals*
*In uniform motion, for regular enough[2] entities, a temporal sequence of relations in a finite time interval is also finite.*

*Proof.* A qualitative representation partitions the phase space of two *regular enough* finite entities in a finite number of regions, i.e., the qualitative relations. Therefore by moving in uniform motion in a finite time interval the system goes through a finite number of such regions, i.e., the resultant temporal sequence of relations must be finite. □

**Proposition 2.** *Finitude of the Stories Set*
*The set of stories in uniform motion, i.e., the stories set, is finite.*

*Proof.* We cannot rigorously prove that the stories set is finite, but Lemma 3 gives an equivalent condition that help us to see that the number of possible stories must be finite in most qualitative representations: if we prove that there is a story with more or an equal number of relations than any other, then the stories set must be finite. This is the case in RCC (Fig. 4), where the longest story is $S_5$. □

**Lemma 3.** *The longest story*
*The stories set is finite, if and only if it exists a longest story, i.e., a story that has more or equal relations than any other.*

# Investigating Representational Dynamics in Problem Solving

Benjamin Angerer
Institute of Cognitive Science
University of Osnabrück
benjamin.angerer@uos.de

Cornell Schreiber
Department of Philosophy
Research Platform Cognitive Science
University of Vienna
cornell.schreiber@univie.ac.at

## Abstract

Successful problem solving relies on the availability of suitable mental representations of the task domain. In more complex, and potentially ill-defined problems, there might be a wide variety of representations to choose from and it might even be beneficial to change them during problem solving. To explore such dynamics on the representational level, we developed a complex spatial transformation and problem solving task. In this task, subjects are asked to repeatedly mentally cross-fold a sheet of paper, and to predict the resulting sheet geometry. Through its deliberate under-specification and difficulty, this task requires subjects to find new and better fitting representations – ranging from visuospatial imagery to symbolic notions. We present an overview of the task domain and discuss various ways of representing the task as well as potential dynamics between them.

## 1 Introduction

Often, the difficulties of problem solving lie not only in how to perform heuristic search, but start with how to *understand* a given task [Van88]. In the study of problem solving, task understanding is typically conceptualised as "setting up" one's mental representation of the problem – its goals, constituents and possible operations – and considered a preparatory phase before the actual problem solving activity ensues in a subsequent solution phase [SH76, Vos06]. However, there is empirical evidence which suggests considerable interaction between task understanding and problem solving. For example, evidence suggests that pertinent phenomena such as insight, analogy, and transfer can be explained best in terms of changes of one's representation *during* problem solving [GW00, KOHR99, KRHM12]. Furthermore, developmental studies have shown that the use of different solution strategies, potentially employing distinct problem representations, might "overlap" during problem solving [Sie02, Sie06]. This suggests that *representational dynamics*, i.e. ongoing changes to how one represents a given task, might play an essential role throughout problem solving.

However, to date systematic investigations of representational dynamics, mapping out problem solving activity on the "representational level", are still missing. Given the predominant focus on the research of heuristic search, tasks are usually designed to constrain subjects to well-defined problem spaces [Goe10]. Considering how closely such problem spaces resemble their task environment, representational dynamics are thus already precluded by task design and presentation. In contrast, we propose the investigation of tasks that elicit representational dynamics as a regular part of problem solving. In order to make progress in solving such a task, subjects are continuously challenged to acquire more knowledge about – and potentially change their perspective towards – the task domain.

To this end, we present a complex spatial transformation and problem solving task in the domain of iterated paper folding. In the following, our focus lies on a first description of the task domain (Sect. 3). With this, we establish a prerequisite for systematic representational-level analyses of problem solving in the targeted domain. As a first step towards such analyses, we conclude with a discussion of various ways of representing the task mentally and the potential dynamics between those (Sect. 4), and a few final remarks discussing further aspects of this task which are beyond the scope of this introductory paper (Sect. 5).

## 2 Task: Iterated Paper Folding

The task we want to discuss in this paper consists of two subtasks which are presented consecutively, i. e. only after completion of the first subtask the second one is revealed:

1. Imagine cross-folding a sheet of paper and inspect the folded sheet,
2. Draw 2D sketches of the forms of edges on each side of the folded paper.

In subsequent iterations of the task, the number of times the sheet is to be cross-folded (and its sides to be sketched) is *incremented* – resulting in multiply-folded sheets ("folds"). Hence, while the subtasks stay the same in all iterations, the complexity of the folds to be made is increased with each iteration.

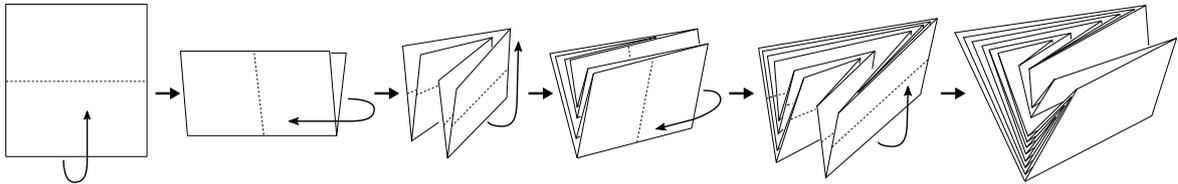

Figure 1: Cross-folding a sheet of paper five times (by alternating between perpendicular folding directions).

This task has several characteristics that make representational dynamics an essential part of human subjects solving it successfully[1]. As a prerequisite for this, it is important to note that the domain of iterated paper folding is such that problem solvers will find a variety of generally distinct representations effective, each with their merits and weaknesses. For instance, the domain can be approached visuospatially (as a case of mental imagery) or analytically (as case of spatial or even symbolic reasoning), but also more intricate combinations of these approaches are feasible (see Sect. 4). Moreover, depending on one's prior knowledge, and the experience gathered in successive iterations, one's conceptual understanding of the domain's underlying principles (see Sect. 3) permits the development of increasingly efficient representations. Taken together, these characteristics provide ample opportunity for representational dynamics.

In the initial engagement with the task, the noted variety of representational possibilities is likely to be most apparent, since the task is posed as an *ill-defined problem* [Ree15, Sim73, Rei65]: Besides the instructions for the first subtask being deliberately vague, no further verbal or graphical hints are provided. Thereby the task instructions do neither "prescribe" the use of specific representations or procedures, nor do they state a precise goal. To form a productive understanding of the task, subjects have to explore the task domain, potentially considering different kinds of more or less suitable representations and procedures [BV02, Goe10]. At the outset, problem solving activity will thus likely be relatively idiosyncratic, taking place in incomplete or incoherent, and dynamically changing problem spaces based on the subject's prior experiences with and familiar knowledge about paper folding.

While at later stages problem solving activity should become substantially more coherent, the reason to still expect sustained dynamics lies with the task's *iterative procedure*. Since with each iteration the complexity of the folds increases substantially, generating them gets progressively more challenging. Thus, simply applying the representations and procedures developed so far to the new challenges will not suffice. To counter these increasing cognitive demands, subjects will instead have to find more efficient ways of representing and manipulating cross-folds – furthering their understanding of the task domain in the process, and effectively changing the problem spaces they are operating in.

---

[1] Presuming that from the start one would apply a form of representation that is well-suited for the complexity of later iterations, and assuming the availability of unlimited processing capacities, it would theoretically be possible to solve our task without changes in how it is represented – but both of these are not the case for typical human subjects (owing to both, the capacity limits of mental imagery, and a lack of detailed knowledge about the consequences of iterated folding).

**Related work**

While the task presented has been newly developed and there is to our knowledge no work directly concerning iterated cross-folding, there is a vast amount of work on other types of paper folding. For instance, there are psychological investigations of the mental folding of cube nets, and unfolding of mutliply-folded sheets with holes punched into them [HHPN13, SF72]. There is also work on the qualitative modelling of these kinds of tasks [Fal16], as well as mathematical and algorithmic descriptions of more complex paper folding such as Origami [DO07, IGT15]. Particularly noteworthy with respect to our work is a study of origami folding tasks which investigated how people reconceptualise the task and its constituents over the course of the study, thus equally emphasising possibilities for conceptual and representational change [TT15].

## 3 Overview of the Task Domain

In the following we provide a systematic overview of the domain of iterated paper folding in terms of its task-relevant entities and their regular interrelation, which subjects might represent in some form or another when engaging with the task. With this, we establish an objective point of reference for conducting and discussing representational-level analyses.

The overview is divided in three parts: First, Sect. 3.1 explains the *procedural details* of how to fold (illustrating the extent of under-specification in the first subtask). Then, Sect. 3.2 describes the *forms* the sheet's sides assume after being folded, and which are closely related to the sketches asked to be drawn for the second subtask. Finally, Sect. 3.3 describes the *relations* holding between these sides and between consecutive folds, respectively.

### 3.1 Folding Procedures

Cross-folding can be defined as *folding a sheet such that it is halved in middle by each fold and such that consecutive creases are perpendicular to each other*. Ignoring the sheet's size, thickness, and aspect ratio – the n-th cross-fold $F_n$ is uniquely determined by:

(a) the initial spatial orientation of $F_0$
(b) the number of times folded ($n$)
(c) the folding procedure used

A folding procedure is determined by how exactly the sheet is being folded, most importantly the *direction* in which the fold is made. Initially (for $F_0 \rightarrow F_1$), there are 8 directions in which we can fold, two of which yielding identical folds (Fig. 2). Since cross-folding requires successive creases to be perpendicular, from $F_1$ onwards there are only 4 directions that can be chosen.[2]

When combining folding directions arbitrarily, there are $2^{2n+1}$ different folding procedures for $F_0 \rightarrow F_n$. Of special interest, however, are procedures in which folding directions are combined *systematically*, such as alternating between the same two perpendicular directions (e.g. Fig. 1). Using such a procedure limits the number of possibilities to 32 (8 directions × 4 perpendicular directions).

Alternatively, a cross-fold can also be achieved by introducing a 90°-rotation-step between two foldings in the same direction (e.g. Fig. 3). For such a procedure, there are only 16 possibilities (8 folding directions × 2 rotation directions).

Even though all of these procedures yield a cross-fold and hence fulfil the first sub-task equally, it is important to distinguish between them. Since the cross-folds produced by them differ in certain respects, they are relevant when trying to identify subjects' representations and evaluating their performance in the second subtask.

---

[2]Whereas folding in one of the other 4 directions would yield a *parallel fold*.

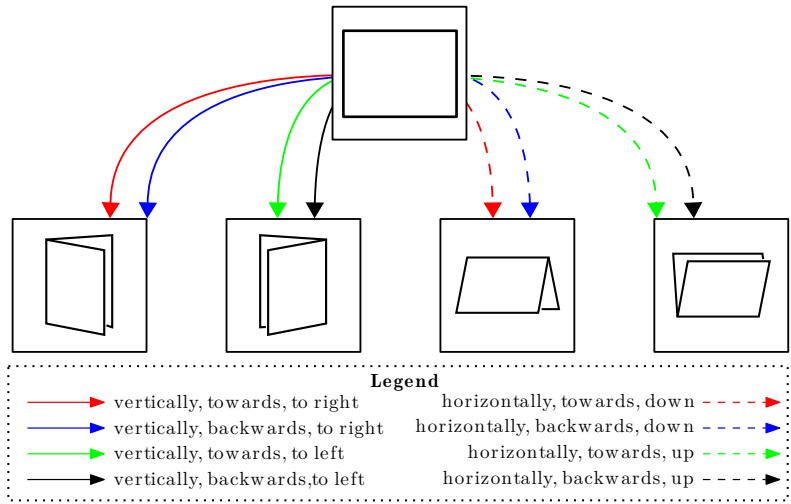

Figure 2: Eight possibilities of turning $F_0$ into $F_1$.

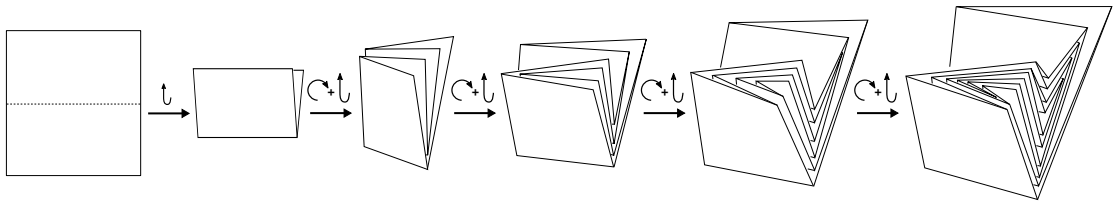

Figure 3: A folding procedure which rotates the sheet 90° clockwise before folding horizontally-towards-up (rotation steps not depicted)

### 3.2 Fold Forms

Describing the consequences of folding on the sheet, we can identify the occurrence of regular schematic forms, which all cross-folds have in common.

Depicting each side of a cross-fold two-dimensionally (as required by the second subtask) five *basic forms* can be observed. They are the forms of all folds from $F_2$ onwards, namely: Two uncreased rectangles, the crease itself ($I$), a single creased edge ($V$), a double V ($DV$), and a nested V ($NV$).

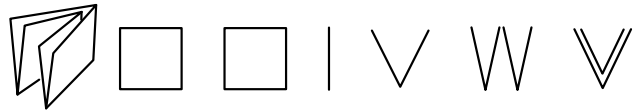

Figure 4: An exemplary illustration of $F_2$, its six sides, and their positions: rectangles (front, back), I (right), V (bottom), DV (left), and NV (top).

The rectangles, $I$, and $V$ (re-)appear unchanged in all folds, but $DV$ and $NV$ only retain their basic shapes, with every $DV_{\geq 2}$ consisting of two sub-figures side-by-side, and every $NV_{\geq 2}$ consisting of two nested sub-figures. As the number of folds increase, they show more complex creasing of edges, hence they are referred to as a fold's two *complex sides*.[3]

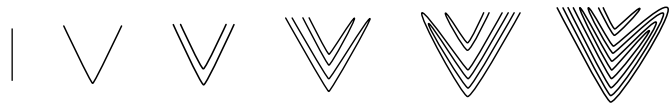

Figure 5: $NV_0$ to $NV_5$ (left to right), as produced by the procedure shown in Fig. 1.

Looking at $F_1$ and $F_2$, it might seem that while there are many different folding procedures (see Sect. 3.1),

---

[3]Hence, the remaining description of our task domain will be mainly concerned with those two sides.

they always bring forth the exact same folds, only in different spatial orientations. Yet, with higher fold numbers certain differences start to appear.

### 3.2.1 Left-handed & Right-handed Folds

Beginning with $F_3$ folds are *chiral*, i.e. they are no longer identical with their mirror images and we have to distinguish left- and right-handed variants. For instance, the folding procedure shown in Figures 1 and 5 produces folds of alternating chirality.

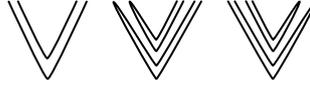

Figure 6: The achiral $NV_2$ next to left- and right-handed variants of $NV_3$.

### 3.2.2 In-folding & Out-folding

Starting with $F_4$, different folding directions produce yet another distinction: Depending on whether a new fold $F_n$ encompasses the $V_{n-1}$ with the $NV_{n-1}$ or the other way around, we distinguish *out-* and *in-folds* (denoted by a superscript I and O). Folding procedures with fixed direction and rotation will always yield out-folds, whereas non-rotating or alternately-rotating procedures can produce both kinds.

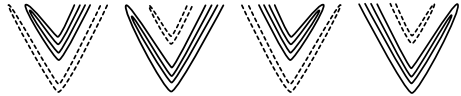

Figure 7: From left to right: Left-handed $NV_4^O$, $NV_4^I$, right-handed $NV_4^O$, $NV_4^I$.

### 3.2.3 Mixed Folds

Furthermore, depending on the folding direction either an in- or an out-fold is produced each time we fold. This means that if allowing arbitrary combinations of folding directions, folds cannot only be "purely" in- or out-folded, but they can be *mixed folds*, with an ever increasing number of possible combinations of in- and out-folding. This results in a total number of $2^{n+1}$ fold variants for $F_n$ (as opposed to an upper bound of 32 variants if we disallowed arbitrary folding directions, and hence mixed folds).

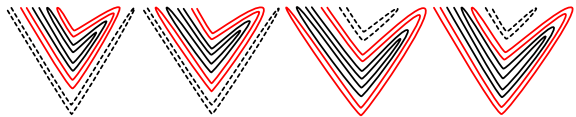

Figure 8: Four different $NV_5$, from left to right: Out-fold, mixed (in-folded out-fold), mixed (out-folded in-fold), in-fold.

## 3.3 Regular Relations

While in theory, every detail of how a fold will look like and how its sides relate to each other can be derived from the chosen folding procedure, these details can be relatively hard to see. Hence, in the following section, we describe the most important of these relations holding in the general case.

### 3.3.1 Within-fold Regularities

While folds can vary in their orientation, and in the detailed structure of their sides etc., the spatial relations holding between a fold's individual sides are the same for all cross-folds (see Table 1).

Table 1: An overview of the spatial relations between the sides of a fold.

|    | V | DV | NV |
|----|---|----|----|
| I  | ⊥ | ‖  | ⊥  |
| V  |   | ⊥  | ‖  |
| DV |   |    | ⊥  |

(‖ = spatially opposite, ⊥ = perpendicularly adjacent)

### 3.3.2 Between-fold Regularities

In an equally general manner, we can describe the relations holding between the four non-rectangular sides of two consecutive folds $F_n$ and $F_{n+1}$ (for $n \geq 2$). Table 2 presents these relations as rules of how to manipulate each side of $F_n$ two-dimensionally in order to generate $F_{n+1}$'s sides from them.

Table 2: An overview of the generative relations between the sides of consecutive folds.

| $I_{n+1}$: | $V_{n+1}$: | $DV_{n+1}$: | $NV_{n+1}$: |
|---|---|---|---|
| new | fold($I_n$) | align($V_n, NV_n$) | fold($DV_n$) |

## 3.4 Summary

While there are many more advanced facts about the task domain[4], above we have introduced the domain's basic properties: (a) The different procedures that can be used to make cross-folds, (b) the forms a sheet's sides assumes after being cross-folded in different ways, and (c) the regular relations holding between a fold's sides.

Following from these basic properties, the large number of possible cross-folding procedures, with its intricate distinctions in higher iterations (chirality, in-/out-folding etc.), as well as the overall increasing complexity of higher folds, present particular challenges to subjects which will potentially give rise to representational dynamics. Additionally, the distinctions presented here are also relevant when trying to identify the representations subjects might use in solving the task.

## 4 Discussion: Representational Dynamics

In the following, we discuss which representational dynamics can be expected in the task. We present how people have been shown to solve mental folding tasks in general, using different varieties of representations. On this basis, we point out potential representational dynamics that can ensue within and between these kinds of representations in the domain of iterated paper folding.

Generally, research on spatial transformation distinguishes between two kinds of solution approaches which presume largely different kinds of representations. As indicated earlier, there are *visuospatial approaches*, i.e. imagining a 3D object, transforming it, and then "seeing" the result, and *analytic approaches*, i.e. understanding and solving the problem based on explicit domain knowledge [HHPN13]. Furthermore, visuospatial and analytic approaches are usually conceived as the poles of a continuum, allowing for *mixed approaches* [GF03].

When faced with the repeated challenges of the task, subjects will have to change their perspectives towards the problem several times, thus furthering their understanding of it, and taking advantage of the merits of different approaches. It is thus necessary to discuss each approach in some detail.

### 4.1 Visuospatial Approaches

In terms of a visuospatial approach, the mental imagery of paper folding is typically conceived of as the mental analogue of folding physically [SF72]. The physical process of folding is a non-rigid transformation, i.e. folding an object affects its individual parts differently [HHPN13]. In cross-folding, for instance, a single act of folding affects a sheet's sides in distinct ways (cf. Table 2). Consequently, in order to mentally form a visuospatial representation of a fold as whole, subjects will likely require multiple repetitions of the same transformation of a sheet, while variably attending to its individual sides.

---

[4]For instance, some readers might be interested that the number of nestings in a fold's $NV_n$ corresponds to the partial sum $\sum_{i=0}^{n-1} 2^{\lfloor \frac{i}{2} \rfloor}$ (sequence A027383 in the OEIS [OEI17]).

The transformation of a visuospatial representation is usually understood as akin to sensorimotor transformations, i.e. an analog transformation of one visuospatial representation into another one mediated by a motor process [Iac11, MK09]. While this transformation is described as analogous to the physical act of folding, visuospatial representations often already leave out many physical details which are irrelevant to the task (such as aspect ratio or size). Crucially, in such a process information on the parts and their spatial interrelation is merely implicit. Presuming a certain everyday familiarity with the activity of cross-folding, visuospatial approaches have the advantage that subjects can perform them without much explicit knowledge about the task domain. Their effectiveness, however, is limited to spatially rather simple or very familiar complex objects [BFS88].

Consequently, when attempting to mentally cross-fold for the first few times, it might be possible to fold $F_1$ and $F_2$ in this manner. However, with increasing complexity of the folds, even attending to their visuospatial representations side-by-side will eventually become too demanding, and subjects have to find other approaches. Consider for example Fig. 9, where a $DV_2$ is being transformed into an $NV_3$ (as part of folding $F_3$ from $F_2$). Depending on one's familiarity with cross-folding, this might already be a very advanced transformation and at the boundary of what a typical subject is able to achieve in a purely visuospatial manner.

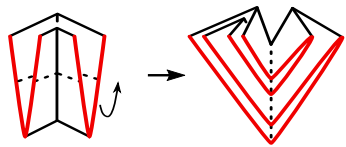

Figure 9: $F_2 \rightarrow F_3$, attending to the transformation of $DV_2$ into $NV_3$.

## 4.2 Analytic Approaches

Analytic approaches to mental folding are more straightforward cases of *problem solving* or *spatial reasoning*. To this end, the mental object is construed as a symbolic representation of interrelated parts. As opposed to visuospatial approaches, spatial information is not implicit in such representations, but has to be made explicit as a semantic relation between symbolic entities. In cross-folding, subjects might represent a fold in terms of the regularities presented in Sect. 3. For instance, $F_2$ can be represented as a nested structure, comprising six schematic 2D forms (cf. Sect. 3.2) which stand in regular spatial relations (cf. Table 1). Additionally, each of those forms can be further decomposed into a spatial configuration of edges. Given such a representation, a successor fold can be realised as a rule-based construction, i.e. a symbol-by-symbol translation according to the between-fold regularities (cf. Table 2).

Crucially, the feasibility of an analytic approach depends on the subject's explicit knowledge of the task domain. Thus, analytic approaches are unlikely to be adopted in the initial phase of the task when subjects are still exploring the ill-defined problem. Only after they have gathered sufficient knowledge, such as the sheet's basic forms and their spatial relations, analytic approaches might start to occur. For example, subjects might utilise their knowledge that some schematic forms are the same for all $F_n$, i.e. the rectangles, $Is$ and $Vs$. Beyond that, the transformation of an $NV$ and $V$ into a $DV$ also lends itself to an analytic solution, since it does not involve complex visuospatial manipulations besides aligning two sides of the previous fold (Fig. 10). While analytic approaches thus allow to avoid otherwise complex operations, the lack of visuospatial representations during problem solving can also lead to paradoxical situations: For instance, aligning two wrong sides would yield a symbolic representation of a physically impossible fold state.

As it gets successively more demanding to represent the sides of higher-numbered folds in terms of simple derivations of the basic forms, ultimately an analytic approach requires an even more economical, *syntactic* way of representing sides. This could be achieved by encoding the number of nestings of edges on the open end of a side. For instance, one could use `0` to encode a simple open edge, `1` for a looped edge, `2` for two nested looped edges etc. So `[0,0]` would represent a single $V$, and `[0,0,0,0,0,0,1,1,4,1]` an $NV_5^O$ (such as on the right of Fig. 11). Based on such a syntax, one could formulate a recursive procedure which – using the sides of $F_2$ as base cases, and the regularities in Table 2 as transformation rules – can generate the complex sides of any $F_n$.

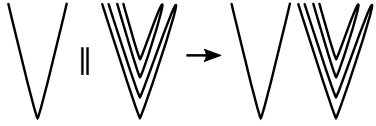

Figure 10: Aligning two existing figures of spatially opposite sides ($V_3$, $NV_3$) to form $DV_4$.

### 4.3 Mixed Approaches

For the most part however, neither visuospatial nor analytic approaches will likely be used exclusively for any longer period of working on the task. We can rather expect both of them being employed, making use of their respective merits, and with various forms of dynamics taking place between them.

A straight-forward case of interaction would be to generate solutions with one of the approaches, but to employ the other one for checking the results. For instance, since the visuospatial transformation illustrated in Fig. 9 is already quite difficult, subjects might be well advised to check their results explicitly against relevant knowledge. On the other hand, a potentially inadequate posit from a symbolic construction, as in the paradoxical case above, can be verified and corrected with the help of visuospatial manipulations.

But there are more intricate forms of dynamics, as well. Notably, explicit knowledge can be utilised variously in order to *scaffold* or *augment* complexity-bounded visuospatial thinking. For example, instead of one holistic visuospatial representation of the sheet, subjects could use visuospatially less demanding 2D representations of each side, while maintaining the correct spatial relations between them explicitly (cf. Table 1). And even within a visuospatial 2D representation of one of the fold's sides, advanced domain knowledge could also allow decomposing a complex side into simpler sub-figures, which by themselves are easier to deal with visuospatially (folded, rotated etc.) once more (Fig. 11).

Finally, the recognition of perceptually analogous features of different folds (requiring visuospatial representations) might lead to the identification and rule-like representation of general features of cross-folding. For instance, noticing the perceptual similarity between successive $NV$s (cf. Fig. 5) might lead to the rule-like hypothesis that *all NV*s are composed of two nested sub-figures.

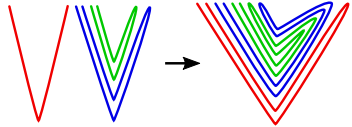

Figure 11: Transforming a $DV_4$ into an $NV_5$, colours marking the different sub-figures.

### 4.4 Summary

Above we have outlined an account of representational dynamics for the proposed task of iterated paper folding. According to this, how subjects approach the task can be expected to vary, dependent on the difficulty of its current stage and the subject's explicit knowledge of the task domain.

While in the initial stages of the ill-defined problem, subjects will likely be successful with visuospatial approaches, in later, more difficult iterations, subjects will profit from changing to more analytic approaches. Therefore, it becomes imperative to gather explicit knowledge of the task domain. However, for the most part subjects will likely follow mixed approaches – as for the relative merits and weaknesses of both visuospatial imagery and knowledge-based constructions. It might thus be more apt to conceive of their development as mutually dependent.

## 5 Final Remarks

We have presented a new spatial transformation and problem solving task of iterated cross-folding and outlined an analysis of the task domain.

Furthermore, we have discussed how subjects can approach the task domain with a variety of representational forms, each of which can be subject to representational dynamics by themselves, but with dynamics also occurring between these varied approaches [SSB13]. However, the systematic task description provided may have given the impression that subjects actually show equally systematic behaviour when approaching the given task. Yet, according to our prior experience with this task such an assumption would be problematic. For instance, we hardly touched upon the varied roles both perceptual and structural cross-domain analogies can play in scaffolding representations [CPS12, Dun01]. In a similar way, metaphors can also play an important role in changing the task domain's conceptualisation [Ami09]. An overall more naturalistic discussion, describing more ephemeral aspects of working on this task (including "task-extrinsic" aspects such as mind-wandering) is provided in [Sch15].

Regarding the theory of problem solving, the extent of representational dynamics observable in a task such as the one presented here might lead us to doubt the notion of stable problem representations in general. Yet, the question of the potential changeability of problem representations has, for example, also lead to several proposed extensions of problem space theory which try to address the problem by the assumption of additional search processes [KBVK14, KD88, SK95].

Ultimately, in order to make progress in these questions and advance theory, we would need wider-ranging representational-level analyses of subjects solving the presented task, and tasks similar to it. We hope that the task domain and first analyses we presented here have set a good starting point for future endeavours of this kind.

### Acknowledgements

We thank Stefan Schneider for his invaluable help during cognitive task analysis.

# An Approach to Compose Shapes Described Qualitatively: A Proof-of-Concept

Albert Pich and Zoe Falomir

## 1 Extended Abstract

This paper presents a proof-of-concept towards solving spatial reasoning tests which deal with object composition. This approach is inspired by the Qualitative Shape Descriptor (QSD) by Falomir et al. (2013), the juxtaposition scheme QSD-Jux by Museros et al. (2011) and the framework Point-Line-Circuit-Area (PLCA) by Takahashi et al. (2015). This new approach, LogC-QSD, uses circuits to describe networks of composed objects and QSD to describe the shape of the boundary of the resulting composition.

LogC-QSD differs from QSD-Jux in the use of circuits to describe networks between the composed objects. LogC-QSD can juxtapose more than two objects using different connections (i.e. only one point connection). And LogC-QSD differs from PLCA by describing of the shape of the objects and the circuit qualitatively. Attributes as angles, lengths, or convexities are not described in PLCA.

As a proof-of-concept, the presented approach has been implemented using Prolog programming language, which is based on Horn clauses and $1^{st}$ order logic. The testing framework has been SWI-Prolog (Wielemaker et al., 2012) and promising results are obtained.

### 1.1 Overview of LogC-QSD

**Target objects**

The target objects of LogC-QSD are those which can be described with QSD, that is, a two-dimensional object with at least three relevant points in its boundary. Thus, a set of relevant points, denoted by $\{P_0, P_1, ..., P_N\}$, $N \geq 3$, determines the shape of the object. Each relevant point $P_i$ is described depending on the relation appearing between $P_i$ and the previous point, $P_{i-1}$, and $P_i$ and the following point, $P_{i+1}$. And each $QSD_i$ is composed by a set of four features, which are defined as:

$$QSD_i(P_{i-1}, P_i, P_{i+1}) = < EC_i, A_i | TC_i, C_i, L_i >$$

where,

- *EC* refers to the Edge Connection, which can be *line_line*, *line_curve*, *curve_line*, *curve_curve* or *curvature_point* depending if the point is connecting lines, curves, or it is a curvature point;

- *A* refers to the Angle at the relevant point and takes the following values *very-acute (va), half-right (hr), acute (a), right (r), obtuse (o), 3-quarters-right (tqr), very-obtuse (vo), plane(pl)*.

- *TC* refers to the Type of Curvature and takes the following values *very_acute, acute, semicircular, plane, very_plane*.

- *C* refers to the convexity at the point $P_i$ and takes the values: *concave* (cv), *convex* (cx) and *plane* (pl).

- *L* refers to the length of the edge between $P_{i-1}$ and $P_i$ and takes the vales: *smaller-short (ss), short (s), larger-short (ls), quarter-longest (ql), smaller-medium (sm), medium (m), larger-medium (lm), half-longest (hl), smaller-long (sl), long(l), larger-long (ll),* and longest (lst).

The QSD descriptions of the objects appearing in the Fig. 1 objects are:

```
?- get_QSD_List(objectA,A).
A = [[p1,line-line,r,cx,m],[p2,line-line,r,cx,m],[p3,line-line,r,cx,m],[p4,line-line,r,cx,m]] .
```

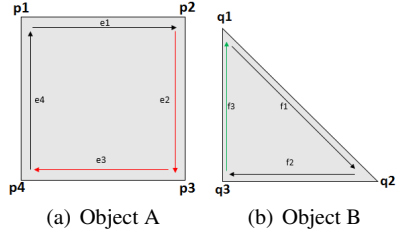

(a) Object A  (b) Object B

**Fig. 1** Examples of objects described by QSD.

That means that objectA has 4 points ($p_1$, $p_2$, $p_3$, $p_4$) which are described using QSD, the description of $p_1$ mean that it connects two lines (line-line), has a right angle (r), the angle defined is convex (cx) and the length is medium (m). The rest of the points have the same description. In the same way, the object B is described.

```
?- get_QSD_List(objectB,B).
B= [[q1,line-line,a,cx,m], [q2,line-line,a,cx,hl], [q3,line-line,r,cx,m]].
```

**Connections between Objects**

The connections are described qualitatively according to their features and cover objects connected by edges or by points. A composition between two objects always have a pair of connections, which are defined as follows:

$$C_i(A,B) = < K_C, P_A|L_A, P_B|L_B, A_C >, i = \{1,2\}$$

where:
- $K_C$ is the kind of connection between the points or lines;
- $P_A|L_A$ is the point or line of the object A involved in the connection;
- $P_B|L_B$ is the point or line of the object B involved in the connection;
- $A_C$ is the exterior angle of the connection that only takes values when the same point belongs to both pair of connections.

Fig. 2 shows some *Connections* as examples: *point-point* (p-p), *point-line* (p-l) and *line-point* (l-p).

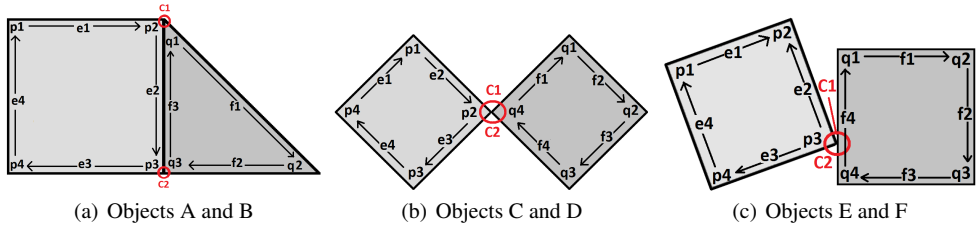

(a) Objects A and B    (b) Objects C and D    (c) Objects E and F

**Fig. 2** Kinds of connections: (a) point-point/point-point; (b) point-point/point-point; (c) point-line/line-point.

Let us name $(A,B), (C,D)$ and $(E,F)$ as the pair of objects in the previous Figure, then their connections can be defined as:

- $C_1(A,B) = (point\_point, p2, q1), C_2(A,B) = (point\_point, q3, p3)$;
- $C_1(C,D) = (point\_point, p2, q4, right), C_2(C,D) = (point\_point, q4, p2, right)$;
- $C_1(E,F) = (point\_line, p3, f4, acute), C_2(E,F) = (line\_point, f4, p3, obtuse)$.

That is, A and B are connected in $C_1$ by points $p_2$ and $q_1$ and also in $C_2$ by points $q_3$ and $p_3$. Then, C and D are connected in $C_1$ by $p_2$ and $q_4$ and also in $C_2$ by $q_4$ and $p_2$, note that even if the same points are involved, two QSD descriptions are required for the same point since it connects different pairs of points, that is, $QSD(p1,[p2,q4],q1) \neq QSD(q3,[q4,p2],p3)$. Furthermore, as the same point belongs to both connections, an exterior angle is required. The objects E and F are connected in $C_1$ by $p_3$ and $f_4$ and also in $C_2$ by $f_4$ and $p_3$, as mentioned before, two descriptions are needed since $QSD(p2,[f4,p3],q1) \neq QSD(q4,[p3,f4],p4)$, as well as an exterior angle.

**Logic Composition of QSD (LogC-QSD)**

A Logic Composition of objects described qualitatively (LogC-QSD) is defined as follows. Given two objects, A and B, described according the Qualitative Shape Descriptor (Falomir et al., 2013) ($QSD_A, QSD_B$) and the connections between them ($C(A,B)$) the QSD description of the composed object is obtained.

$$LogC(QSD_A, QSD_B, C(A,B)) = QSD_{AB}$$

The result is another QSD description, $QSD_{AB}$, that can be composed again. In general, the LogC-QSD can compose N objects, requiring the composition of two objects at a time.

The LogC-QSD concerns two steps. In the fist one, a network describing the composition of objects is built. In the second step the network is characterized regarding: (i) the QSD of the provided objects, (ii) the kind of connections, and (iii) the new composed features of angle, convexity and length in the object connection.

**Object Composition Network**

This network is built by the edges of the objects and the kind of connections between them. Depending on the kind of connection happening between the objects, a different procedure is followed. As QSD describes the objects by their points in a clockwise order, the LogC-QSD also follow this convention. The network is constructed according to the next steps:

1. Starting from the first object, A, from the first relevant point to the point or line involved in the first connection, all the edges are stored to the resulting network $QSD_N$;
2. When a connection happens, the kind of connection and the pair of edges involved are stored together. In general, if a point ($p_i$) is involved in the connection the following edge ($e_i = (p_i, p_{i+1})$) is stored. Otherwise, if a line is involved, the edge which contains this line is stored;
3. The rest of the edges from the next object, B, are stored in $QSD_N$ until the next connection is reached;
4. The step 2 is repeated in the opposite order, that is, from the second object, B, to the first object, A;
5. Finally, the rest of the edges from the first object are stored in $QSD_N$.

As an example, the construction of the composition network of objects in Fig. 3 is presented. Note that this network is constructed progressively taking the objects in pairs as it is explained next.

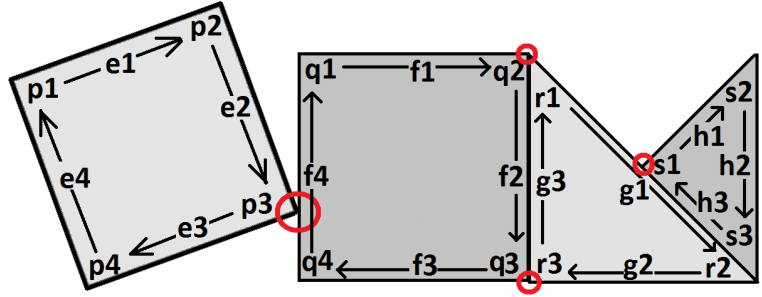

**Fig. 3** Composition of the objects A,B,C and D from left to right.

```
?- network(A,p3,B,f4,NAB).
NAB = [e1, e2, [p-l,e3, f4], f1, f2, f3, [l-p,f4,e3], e3, e4].
```

When calculating the composition network between the first object in Fig. 3, A, and the second object, B, the results show that: the edges $e_1$ and $e_2$ are stored in the network ($QSD_{NAB}$), then the point $p_3$ is reached, which belongs to the point-line (p-l) connection, so the kind of connection together with the following edge to the point involved in the connection ($e_3$) and the edge which contains the line ($f_4$) are stored, that is, [p-l,$e_3$,$f_4$]. Then, the rest of the object B is stored until the line involved in the connection ($f4$) is reached, and $f_1$, $f_2$ and $f_3$ are stored. Then, a line-point (l-p) connection is happening, so, following the same process as before, the kind of connection together with the edge which contains the line involved, and the following edge to the point involved ($e_3$) are stored, that is, [l-p,$f_4$,$e_3$]. Finally, the rest of the edges ($e_3, e_4$) from the object A are stored.

```
?- network(NAB,f2,C,g3,NABC).
NXY = [e1, e2, [p-l,e3, f4], f1, [p-p,f2,g1], g2, [p-p,g3,f2], f3, [l-p,f4,e3], e3, e4].
```

The same process is followed in the composition between the third object in Fig. 3, C, and the previously obtained composition network ($QSD_{NAB}$). That is, $e_1, e_2, [p-l, e_3, f_4], f_1$ are stored, then a point-point (p-p) connection is reached and stored together with the corresponding edges, $[p-p, f_2, g_1]$, then the rest of the object C ($g_2$) is stored until the second connection $[p-p, g_3, f_2]$ is reached and stored, and finally the rest of the previously composed network ($f3, [l-p, f_4, e_3], e_3, e_4$) is stored.

```
?- network(NABC,g1,D,s1,NABCD).
NABCD = [e1, e2,[p-l,e3, f4], f1, [p-p,f2,g1],[l-p,g1,h1],h2,[p-p,h3,g2],[p-p,g3,f3],
[l-p,f4,e3],e3,e4].
```

Finally, the composition network between the fourth object in Fig. 3, D, and the previous composition network ($QSD_{NABC}$) is constructed in the same way.

**Describing the QSD of the Object Composition Network**

In this second step, the obtained network is characterized regarding: (i) the QSD of the provided objects, (ii) the kind of connections, and (iii) the new composed features (i.e. edge-connection, angle, convexity and length) defined for the connections.

For any edge that does not take part in a connection, the source point of the edge is chosen and its QSD description is obtained from the knowledge base. For example, given $e_1 = (p_1, p_2)$, the characterization of $p_1$ is obtained:

```
?- edge(e1,PSource,_), hasQSDpoint(objectA,PSource,_,_,EC,A,C,L).
PSource=p1, EC=line-line, A=r, C=cx, L=l.
```

The QSD corresponding to the point $p_1$ is obtained: *[line-line,right,convex,a]*

For the edges with connections, the following distinctions are made:

- If the two connections define a slope. Then, the important feature to analyze is the length:

  – If the connected edges have the same lenght, the composition operation finishes composing the angle and convexity at the connection points.
  – If the connected edges have different length, then these lengths must be added or subtracted, depending on the case, and the angles must be also calculated.

- If the same point belongs to the pair of connections between two objects, an external *Angle* is needed, otherwise infinite possibilities can be found.

For the connection pair line-point [l-p,$g_1$,$h_1$], the endpoint of $g_1$ and the source point of $h_1$ are taken, obtaining $r_2$ and $s_1$ respectively.

For example, for the connection pair point-point [p-p,$h_3$,$g_2$], the source point of the edges are chosen and its QSD description is stored, obtaining:

```
?- hasQSDpoint(objectD,s3,_,_,TC,Angle,Convexity,Length).
PSource=s3, TC=line-line, Angle=hr, Convexity=cx, Length=l.
?- hasQSDpoint(objectC,r2,_,_,TC,Angle,Convexity,Length).
PSource=r2, TC=line-line, Angle=hr, Convexity=cx, Length=lst.
```

Then, the features *edges connected*, *angles* and *convexities* are composed according to the composition tables. In this case, the *length* is taken from the point $s_3$.

```
?- compose_ec(line-line,line-line,EC).
EC=line-line.
?- compose_angles(hr,cx,hr,cx,Angle,Convexity).
Angle=r, Convexity=cx.
```

Therefore, the QSD of the composed point $(s_3, r_2) = [line\text{-}line, right, convex, a]$ is obtained.

For the connection pair line-point [l-p,$g_1$,$h_1$], the endpoint of $g_1$ and the source point of $h_1$ are taken, obtaining $r_2$ and $s_1$ respectively.

```
?- hasQSDpoint(objectC,r2,_,_,EC,A,C,L).
PSource=r2, EC=line-line, A=hr, C=cx, L=lst.
?- hasQSDpoint(objectD,s1,_,_,EC,A,C,L).
PSource=s3, EC=line-line, A=r, C=cx, L=hl.
```

The feature *edges connected* (EC) is operated as before, the *angles* and *convexities* from $s_1$ are composed with the *plane* angle, without requiring an exterior angle, and the lengths are composed, requiring a subtraction/decomposition between $r_2$ and $s_1$. Note that the same composition tables are used for the subtraction/decomposition operation, only changing the order of the variables.

```
?- compose_ec(line-line,line-line,EC)
EC=line-line.
?- compose_angles(r,cx,pl,pl,A,C).
A=r, C=cv.
?- compose_length(hl,L,lst).
L=hl.
```

And the QSD of the composed point $(g_1, s_1)$ is defined as [*line-line*, *right*, *concave*, *hb*].
After characterizing the network, the resulting $QSD_R$ description is obtained as follows.

```
?- features_Points(NABCD,FP).
FP = [[p1,line-line,r,cx,l], [p2,line-line,r,cx,l], [[p3,f4],line-line,v-a,cv,l],
[q1,r,cx,hl], [[q2,r1],tqr,cx,l], [[g1,s1],r,cv,hl], [s2,hr,cx,hl], [[s3,r2],r,cx,l],
[[r3,q3],pl,pl,l], [q4,r,cx,l], [[f4,p3],a,cv,s], [p4,r,cx,l]].
```

Finally, the points with *plane* angles are removed and their lengths are composed. The result obtained is the following:

```
?- remove_Planes(FP,QSD).
QSD = [[p1,line-line,r,cx,a], [p2,line-line,right,cx,a], [[p3,f4],line-line,v-a,cv,a],
[q1,r,cx,hb], [[q2,r1],tqr,cx,a], [[g1,s1],r,cv,hb], [s2,hr,cx,hb], [[s3,r2],r,cx,a],
[q4,r,cx,b_a], [[f4,p3],a,cv,c], [p4,r,cx,a]].
```

When some connections are not specified, the LogC-QSD approach return all the possibles compositions that satisfy those who are specified. Likewise, if no connection is specified, the LogC-QSD approach obtains all the possible compositions between the objects. The application is also able to discard impossible compositions and offers feedback about the reason of discarding.

**Motivation and Future Work**

The motivation of the LogC-QSD approach is to develop algorithms of spatial reasoning able to solve object composition tasks cognitively. Traditional algorithms in AI solve puzzles using heuristics and lots of computational effort since they find the correct solution. As this is not a cognitive solution, that is, it is not usually applied by humans, this paper explores how spatial logics can improve common reasoning about space. Furthermore, since the target objects of the LogC-QSD are described qualitatively and the implementation is based on logics, the LogC-QSD is able to provide feedback that can be understood by humans.

As future work, the LogC-QSD approach intends to improve spatial logic reasoning in intelligent systems (i.e. video game, robotic agents, etc.).

**Acknowledgements** The support by the project *Cognitive Qualitative Descriptions and Applications* (CogQDA) funded by the Central Research Development Fund (CRDF) at University of Bremen through the 04-Independent Projects for Postdocs action and the European Erasmus+ Internship program are very acknowledged.

# Problem-solving, Creativity and Spatial Reasoning: A ProSocrates 2017 Discussion

Ana-Maria Olteţeanu
amoodu@informatik.uni-bremen.de

Zoe Falomir
zfalomir@uni-bremen.de

University of Bremen, Bremen Spatial Cognition Center (BSCC)
Bremen, Germany

## Abstract

The ProSocrates 2017 Symposium aims to blend the topics of problem-solving, creativity and spatial reasoning. In this paper we will discuss the scientific interest of the topic blend. We will then describe the presented works, how they fit the Symposium theme and statistics about the 2nd Edition of ProSocrates. Finally, we will discuss how the topic mix looked like this year, possible future directions and an outlook for how the topics could be combined in the future.

## 1 Introduction

Problem-solving [Newell and Simon, 1972], human creative cognition [Finke et al., 1992, Boden, 2003], spatial cognition [Freksa, 2015], Qualitative modelling [Forbus, 2011] and computational creativity [Colton and Wiggins, 2012] are topics often treated separately, despite their major potential for synergies.

The ProSocrates Symposium aims to blend these topics and observe their interrelations. The ProSocrates Symposium had its first edition at the KogWis 2016 - Space and Cognition conference, and is now at its second edition, which was supported by the Hanse-Wissenschafstkolleg (HWK), Institute for Advanced Study[1].

This paper discusses the scientific interest and topics of ProSocrates (2), and summarizes the research papers and invited talks that were the content of ProSocrates 2017 (3). The way these papers represent the ProSocrates topics of interest and their interrelations this year is discussed in section 4.

## 2 Scientific interest and Topics

**Problem-solving** has been approached in different ways by AI and the study of human cognition. The ability to flexibly solve novel problems with little training is a fundamental component in human intelligence. For example, rescheduling a trip due to unexpected circumstances, playing Tangram/origami or finding a solution to an ill-structured problem [Newell and Simon, 1972] are well known tasks in human problem-solving. Commonsense reasoning, model building and the ability to creatively solve novel problems have an important role in the challenge of approaching general artificial intelligence.

**Computational creativity** focuses on building creative artificial systems capable of creative feats similar to those achieved by humans [Colton and Wiggins, 2012]. Through its applied nature, computational creativity is a great way of studying the type of algorithms and system implementations that can be used to approach artificially creative results, however the processes and representations in the field are rarely compared to those used by humans.

---

[1] http://www.h-w-k.de/

**Human creative cognition** investigates the way humans solve a multiplicity of creative tasks [Mednick and Mednick, 1971], from the simple (coming up with an alternative use for an object) to the complex (solving insight problems)[Batchelder and Alexander, 2012], asking questions about process. However, no cognitive modeling universal set of tools exists for the pursuit of implementing computational approaches to test hypotheses in a unified manner.

**Spatial cognition and reasoning** [Freksa, 2015] is also known to contribute to the development of abstract thought, and to have a role in problem solving. Spatial cognition studies have shown that there is a strong link between success in Science, Technology, Engineering and Math (STEM) disciplines and spatial abilities [Newcombe, 2010]. Qualitative modeling [Forbus, 2011] concerns the representations and reasoning that people use to understand continuous aspects of the world. Qualitative representations are also thought to be closer to the cognitive domain, as showed by models for object sketch recognition [Forbus et al., 2011], for solving Raven's Progressive Matrices intelligence test [Lovett and Forbus, 2017], for solving oddity tasks [Lovett and Forbus, 2011], for 3D perspective descriptions matching [Falomir, 2015] and for paper folding reasoning [Falomir, 2016]. In the context of creativity, spatial descriptors and qualitative shape and colour descriptors and their similarity formulations were tools for object replacement and object composition in the theoretical approach presented by [Olteţeanu and Falomir, 2016] to solve Alternative Uses Test.

Cognitive Systems is a great intersection point for all these topics, bringing together the perspectives of human cognition, cognitive modeling and cognitive artificial intelligence. The interactions between (i) problem-solving and creativity in the context of cognitive systems [Olteţeanu, 2014, Olteţeanu and Falomir, 2015, Olteţeanu and Falomir, 2016, Olteţeanu et al., 2015, Olteţeanu, 2016, Olteţeanu, 2016, Olteţeanu et al., 2017] and between (ii) qualitative modelling and spatial cognition [Falomir, 2015, Falomir, 2016] were studied by the proposed guest editors in the last few years.

The focus of the ProSocrates Symposium and of the ensuing Special Issue is to bring the previous described disciplines together, by bringing in dialogue specialists from each of the fields. Authors of experimental, theoretical and computational work which combine perspectives from at least two of these 4 topics (problem-solving, spatial cognition/reasoning, cognitive systems and creativity) will be invited to submit contributions. The larger aim of integrating these topics is to produce theoretical tools, approaches and methodologies for creative and spatial problem solving in cognitive systems, in a manner that would benefit from such interdisciplinary bootstrapping.

## 3 Summary of works presented at ProSocrates'17

These ProSocrates'17 proceedings contains 6 accepted papers that were presented at the symposium. Each submitted paper was reviewed by two/three program committee members. Moreover, there were 6 invited talks whose abstracts are also included in these proceedings. A very short summary of these contributions follows, with the aim of observing the various topic blends in the next section.

### 3.1 Invited Talks

In his talk, *Reasoning at a Distance by Way of Conceptual Metaphors and Blends*, **Marco Schorlemmer** presented a mathematical model for expressing conceptual metaphor and blending; he then talked about modelling these in combination with the idea of "reasoning at a distance", from Barwise-Seligman theory of information flow.

In the talk *Symbolic models and computational properties of constructive reasoning in cognition and creativity*, **Tarek Besold** presented general principles of computational analogy, with a closer look at HDTP and cognitive plausibility.

In her talk, *Creating and rating harmonic colour palettes for a given style*, **Lledó Museros** presented a qualitative colour theory, the operations to create harmonic colour palettes and their classification as lifestyle.

**Ken Forbus** discussed ideas for using the Companion cognitive architecture to create software collaborators that support creative work, in his talk *Creative Support Companions: Some Ideas*.

In the talk *Modeling visual problem-solving as analogical reasoning*, **Andrew Lovett** presented a computational model that uses analogical reasoning for visual problem-solving (i.e. Ravens Progressive Matrices test).

**Bipin Indurkhya** gave a talk titled *Thinking Like A Child: The Role of Surface Similarities in Stimulating Creativity*. In his talk, he presented examples of puzzles, research on creative problem solving, and two of recent empirical studies to demonstrate how surface similarities can stimulate creative thinking and to designing creativity-support systems.

### 3.2 Research Papers and talks

In the paper *Towards finer-grained interaction with a Poetry Generator*, **Hugo Gonçalo Oliveira et al.** reported on a recent effort towards providing alternative ways of using PoeTryMe to meet user suggestions, including a co-creative interface.

In the paper *Towards the Recognition of Sketches by Learning Common Qualitative Representations*, **Wael Zakaria et. al** reported on building a hybrid technique, in which aspects of machine learning, computer vision, and qualitative representations are mixed to produce a classifier for sketches of four types of objects.

**Juan Purcalla Arrufi and Alexandra Kirsch** submitted the paper *Using Stories to Create Qualitative Representations of Motion*. In this, they described a method to create new qualitative representations of motion from any qualitative spatial representation by using a story-based approach.

In the paper titled *Investigating Representational Dynamics in Problem Solving*, **Benjamin Angerer and Cornell Schreiber** presented a new spatial transformation and problem solving task of iterated cross-folding and outline an analysis of the task domain. In this task, subjects are asked to repeatedly mentally cross-fold a sheet of paper, and to predict the resulting sheet geometry. The authors discuss how subjects can approach the task domain with a variety of representational forms.

In the paper titled *An Approach to Compose Shapes Described Qualitatively: A Proof-of-Concept*, **Albert Pich and Zoe Falomir** presented a proof-of-concept towards solving spatial reasoning tests which deal with object composition, such as those used in measuring human spatial skills. This approach describes the connections between the objects and the shape of the final provided composition.

Two further talks were presented by the Symposium organizers, on state of the art of their research.

**Ana-Maria Olteţeanu** described how computational creative problem solving systems can be used to generate creativity test queries, in her talk *Generating queries with multi-parameter control for creativity tests*.

**Zoe Falomir** in her talk *3D Perspective Reasoning and Paper Folding Reasoning using Computer Games* explained how qualitative models can reason to solve spatial tests (i.e. perspective taking in object description or paper folding) and how the logics behind these models can be embedded in computer games to provide feedback to players so that they can: (i) understand the space transformations involved in the tests and (ii) practice to improve their spatial cognition skills.

## 4 Discussion - topic mix and future outlook

In this ProSocrates edition, all three of the main topics, Problem solving, Creativity and Spatial reasoning were present, together with some interdisciplinary research at the intersection of various pairs. Here is how the various talks and research papers represented the fields.

The talks by Forcus and Besold, with their focus on creative companions and computational analogy, represented the Creativity topic. The papers of Purcalla Arrufi&Kirsch and Zakaria et al represented mostly the topic of Spatial Reasoning.

The rest of the talks and papers were interdisciplinary, representing some intersection of two of the fields. Indurkhya, Anger&Schreiber and Olteţeanu presented talks at the intersection of Problem Solving and Creativity. Lovett, Falomir and Pich represented the Problem Solving and Spatial Reasoning intersection. The talks by Museros and Schorlemmer situated themselves at the Spatial Reasoning - Creativity intersection. This outline of the content of talks for ProSocrates 2017 can be observed in the Venn diagram in Figure 1.

The fact that ProSocrates 2017 managed to promote and attract interdisciplinary papers between the three fields counts as a success for the organizers. We hope this trend to continue for the next edition, and perhaps ProSocrates 2018 will also start seeing some points in the middle of the Venn diagram, at the intersection between all three fields.

## 5 Epilog

The main focus of ProSocrates 2017 and the presented papers is to promote research in and between three possibly synergistic fields. This aims to encourage new research and collaborations, at the intersections of three fields which are very important for the general society.

The guest editors of these proceedings would like to specially thank funding from the Hanse-Wissenschaftskolleg (HWK), Institute of Advanced Study[2]. We are also grateful to the reviewers for their effort in evaluating the papers considered, and for giving highly constructive feedback to the authors. Finally, they also thank CEUR-WS.org for indexing these proceedings.

---
[2]HWK: http://www.h-w-k.de/

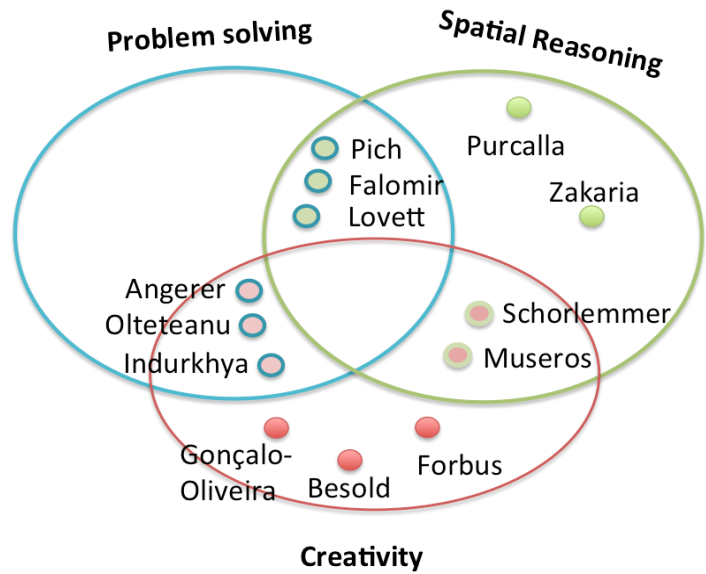

Figure 1: Outline of the contents of the talks distributed in a Venn diagram including Problem-Solving, Spatial Reasoning and Creativity.